\pdfoutput=1
\documentclass{article}

\usepackage[preprint, nonatbib]{neurips_2021}

\usepackage{amsmath,amsfonts,bm,amssymb,amsthm}
\usepackage[all]{xy}

\usepackage{algorithm}
\usepackage{algorithmic} 

\newtheorem{Theorem}{Theorem}
\newtheorem*{Theorem*}{Theorem}
\newtheorem{Lemma}{Lemma}
\newtheorem*{Lemma*}{Lemma}
\newtheorem{definition}{Definition}[section]
\newtheorem{Property}{Property}
 
\def\1{\bm{1}}

\newcommand{\norm}[1]{\left|#1\right|}

\def\sS{{\mathbb{S}}}

\def\sZ{{\mathbb{Z}}}

\DeclareMathOperator{\Ker}{Ker}
\DeclareMathOperator{\Img}{Im}
\usepackage[square,sort,comma,numbers]{natbib}
\usepackage{graphicx}
\usepackage[export]{adjustbox}
\usepackage{float}
\usepackage{tabularx, booktabs} 
\usepackage{multirow}

\usepackage[utf8]{inputenc} 
\usepackage[T1]{fontenc}    
\usepackage{url}            
\usepackage{booktabs}       
\usepackage{amsfonts}       
\usepackage{nicefrac}       
\usepackage{microtype}      
\usepackage{xcolor}         

\makeatletter
\newcommand{\discardpages}[1]{
  \xdef\discard@pages{#1}
  \AtBeginShipout{
    \renewcommand*{\do}[1]{
      \ifnum\value{page}=##1\relax%
        \AtBeginShipoutDiscard
        \gdef\do####1{}
      \fi%
    }%
    \expandafter\docsvlist\expandafter{\discard@pages}
  }%
}
\newif\ifkeeppage
\newcommand{\keeppages}[1]{
  \xdef\keep@pages{#1}
  \AtBeginShipout{
    \keeppagefalse%
    \renewcommand*{\do}[1]{
      \ifnum\value{page}=##1\relax%
        \keeppagetrue
        \gdef\do####1{}
      \fi%
    }%
    \expandafter\docsvlist\expandafter{\keep@pages}
    \ifkeeppage\else\AtBeginShipoutDiscard\fi
  }%
}
\makeatother

\title{Dive into Layers: Neural Network Capacity Bounding using Algebraic Geometry}

\author{%
   Ji Yang$^{*}$ \\
   Enflame-tech \\
    Enflame-tech, 61 ShengXia Road, Pudong New District, Shanghai \\
   \texttt{erik.yang@enflame-tech.com} \\
   \AND
   Lu Sang$^{*}$\\  
   Technical University of Munich \\
   Boltzmannstrasse 3, 85748 Garching, Germany \\
   \texttt{sang@in.tum.de} \\
   \AND 
   Daniel Cremers \\
   Technical University of Munich \\
   Boltzmannstrasse 3, 85748 Garching, Germany \\
   \texttt{cremers@in.tum.de}
  
}

\begin{document}

\maketitle
\let\thefootnote\relax\footnote{$^{*}$ These authors contributed equally.}
\begin{abstract}

The empirical results suggest that the learnability of a neural network is directly related to its size. To mathematically prove this, we borrow a tool in topological algebra: Betti numbers to measure the topological geometric complexity of input data and the neural network. By characterizing the expressive capacity of a neural network with its topological complexity, we conduct a thorough analysis and show that the network's expressive capacity is limited by the scale of its layers. Further, we derive the upper bounds of the Betti numbers on each layer within the network. As a result, the problem of architecture selection of a neural network is transformed to determining the scale of the network that can represent the input data complexity.  With the presented results, the architecture selection of a fully connected network boils down to choosing a suitable size of the network such that it equips the Betti numbers that are not smaller than the Betti numbers of the input data.  We perform the experiments on a real-world dataset MNIST and the results verify our analysis and conclusion. The demo code is publicly available\footnote{\url{https://github.com/Sangluisme/NeuralNetworkBettiNumber}}. 
\end{abstract}

\section{Introduction}\label{sec:into}
Neural networks have rapidly become one of the most popular tools for solving challenging problems (\cite{bengio2009,liu2017}) in various domains such as image understanding~\cite{guo2016deep,maninis2016}, natural language processing~\cite{indurkhya2010, amiajnl2011}, and speech recognition~\cite{deng2013}. In many cases, a major difficulty is to choose an appropriate size of the network that optimally balances the training cost and the expression structure in the data. Empirical results suggest that, for example, fully connected layers, i.e., dense layers with more hidden units, perform better than smaller layers with the same batches of inputs. However, too large networks can lead to excessive computational overhead or overfitting. \par
In computer vision, some works such as~\cite{alom2019, Elsken2019, wistuba2019} use Neural Architecture Search (NAS) to determine the architecture of a network, since they treat architecture selection as a compositional hyperparameter. The advantage of their methods is that they can deal with different types of layers, such as convolution or pooling layer. Some other works ~\cite{simonyan2015, szegedy2014} try to improve initial architecture selection. However, their results are difficult to interpret beyond empirical optimality. Despite the success of these approaches, there is still no general principle for architecture selection.\par
In this paper, we exploit a topological invariant, i.e., Betti numbers, to describe the expressiveness of a neural network architecture given the input data, which is formally introduced in Section~\ref{sec:bg}. Once Betti numbers are used to describe the topological geometric complexity of a structure, a network with efficient expressiveness must represent the topological complexity of the input data, in other words, \emph{the Betti numbers at each layer can not be smaller than the Betti numbers of the input data}. As we will introduce in Section~\ref{sec:main}, the upper bounds on the Betti numbers of each layer are determined by the number of hidden units and layer numbers. Thus, a relation is established between the scales of a network and the topological complexity of the input data. Since the Betti numbers of the input data can be precomputed, our results give a clear indication of the relationship between the size of dense layers and their expression capabilities.
\subsection{Our Contribution}
In this paper, we focus on fully connected networks or multilayer perceptrons (MLPs) with \emph{polynomial} or \emph{ReLU} activation, which are widely used (\cite{boss2020, Ma2017, mildenhall2020}). The impact of the layer size on the performance is analyzed. A fully connected layer can be interpreted as an operator that maps the input into a quotient space (equivalent classes). Given a neural network for classification tasks, a successfully trained network means that the input is mapped to the correct classes. We traverse back to each layer to examine the pre-images of the output classes, and these pre-images are sequences of sets endowed with certain topological properties, these specific topological structures allow us to bound their Betti numbers. a) We use Betti numbers as a quantitative measure of the topological complexity of a network consists of dense layers with ReLU or polynomial activations. b) We derive the upper bounds on the Betti numbers of each layer and show that the architecture selection problem can be characterized by endowing the "efficient topological capacity" of each layer. c) Our results provide mathematical insight into the performance of a network and its capacity. d) In addition, we verify our theoretical results using the MNIST\cite{lecun2010} dataset to show that understanding topological complexity is beneficial in determining the structure of the neural network.

\section{Background}\label{sec:bg}
\subsection{Topology and Algebraic Topology}
Topology, used in this paper refers mainly to topological geometry. Topology is a mathematical branch that studies the properties of objects that are preserved by deformation, twisting, rotation, and extension. From a topological point of view, two objects (or spaces) $X$ and $Y$ are said to be equivalent if there is a continuous function $f: X\to Y$ which has a continuous inverse $f^{-1}$. $X$ and $Y$ are called homeomorphic and $f$ is their homeomorphism. The advantage of using topological properties to characterize data or network structures is that these properties are invariant to some irrelevant features such as scaling, rotation, translation, etc. Using topological complexity to describe the capacity of a network provides an "outline structure" of the output domain.\par
Algebraic topology studies the intrinsic qualitative aspects of spatial objects. It uses algebraic concepts such as groups and rings to represent topological structures. Algebraic topology provides a great tool to analyze and compute the structures of spaces. Simply speaking, algebraic topology studies the' hole' structures of a space. The complexity of the topological geometry is then characterized by the number of 'holes' in different dimensions that space contains. Therefore, a space is divided into simplex cells into different dimensions. Simplex cells are generalized triangles that can be considered as the basis of a topological space. A space is formed by gluing cells together in a particular way. These "glued together" cells form simplical complex. Intuitively, a simplex can be thought as a discretization of a space into triangles that are glued together. After using algebraic topology to assign groups to these cells, the homology tool can be used.
\subsection{Homology and Persistent Homology}
\paragraph{Homology-}Homology groups are mainly used in this paper. Informally, the $n$-th homology group of a topological space $X$, denoted $H_n(X)$ describes the number of $n$-th dimensional "holes" in $X$, the $0-$dimensional hole being the connected component of the space. More strictly, for a topological space $X$, one first constructs a chain complex $C(X)$ that encodes information about $X$. Usually, this is a sequence of abelian groups $C_0, C_1, \dots$, connected by the function $\delta_i: C_i \to C_{i-1}$ which preserves the group structure, the functions are group homomorphisms called boundary operators. The chain complex can be constructed by continuous mappings from the simplex cell to $X$. Then the homology group is defined as $H_n(X) = \Ker\delta_n/\Img \delta_{n+1}$, i.e., the quotient group of kernel of $\delta_n$ and image of $\delta_{n+1}$. $C(X)$ reflects the structure of $X$.
\paragraph{Betti Number-}The $n$-th Betti number $b_n$ of the set $X$ is the rank of the homology group $H_n(X)$, which specifies the maximum number of cuts that can be made before a surface is decomposed into an $n$-th dimensional simplex. Geometrically, the $i$-th Betti number refers to the $i$-th dimensional holes in space $X$. For example, for a circle $\sS^1$, the $i$-th homology group $H_i(\sS^1)$ is zero if $i$ is greater than $1$, since there is no higher dimensional hole; $H_i(\sS^1) = \sZ$ for $i = 0, 1$ and $b_0(\sS^1) = b_1(\sS^1) = 1$ because $\sS^1$ has a connected component and a 1-dimensional hole; for more details see~\cite{Hatcher2000}. Similarly, an $n$-dimensional sphere $H_i(\sS^n)$ is nontrivial if and only if $i$ is $0$ or $n$, i.e. $H_i(\sS^n) = \sZ$ for $i=0$ and $n$, since the $n$-dimensional sphere $\mathbb{S}^n$ has a connected component and an $n$-dimensional hole.
\paragraph{Persistent Homology and TDA-}To compute Betti numbers for arbitrary topological spaces, we borrow the tool of persistent homology. Persistent homology computes topological properties of a space that are invariant to a particular choice of parameters, e.g., a torus has two 1-D holes ($b_0 = 1, b_1 = 2, b_2 = 0, \cdots$), regardless of the inner or outer radius, its orientation or position. The main procedure for finding the persistent homology in a space with simplicial complex structure is to find a real-valued function $f$ satisfying some certain properties on these complices $K$, so that we can form a filtration of the level set $K_i$ ($K(a) = f^{-1}(-\infty,a]$), i.e., $\emptyset = K_0 \subset K_1 \subset \dots \subset K_n = K$. 
Using the properties of filtration, for each dimension $n$ we can find the homomorphism on the simplicial homology groups $H_p(K_i)$ to $H_p(K_j)$, the $p$-th persistent homology groups are the images of these homomorphisms, and the $n$-th persistent Betti numbers $b_n^{i,j}$ are rank of these groups. With the persistent homology properties, we are allowed to represent the persistent homology with \emph{barcode} or \emph{persistent diagram}. \par
\begin{figure*}[!ht]
  \centering
  \setlength\tabcolsep{2pt} 
  \def\arraystretch{0} 
  \newcommand{\mywidth}{0.44\textwidth} 
  \newcommand{\mywidthlr}{0.25\textwidth} 
  \newcommand{\mywidths}{0.45\textwidth}
  \newcolumntype{Y}{>{\centering\arraybackslash}m{\mywidths}}
\begin{tabular}{YY}
\includegraphics[width = \mywidths]{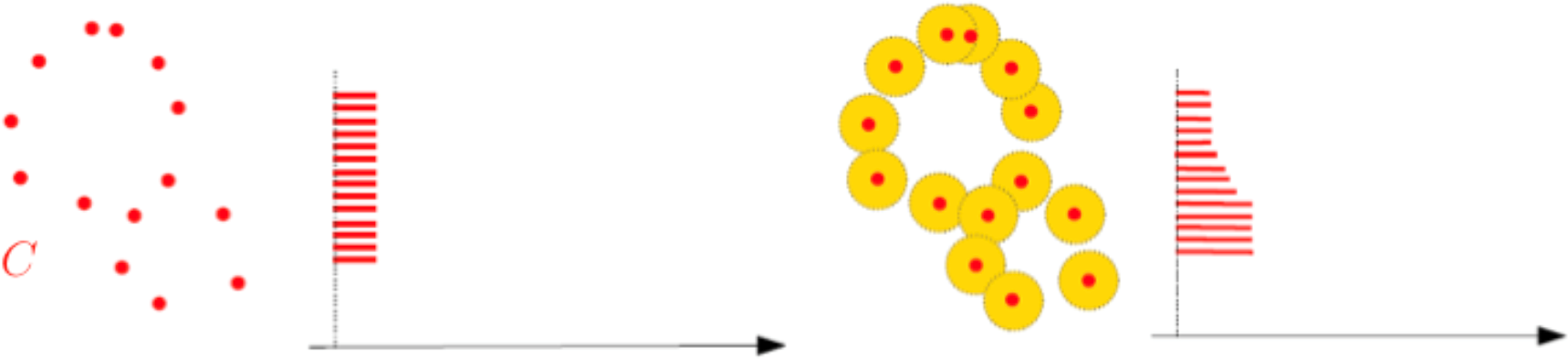} &
\includegraphics[width = \mywidths]{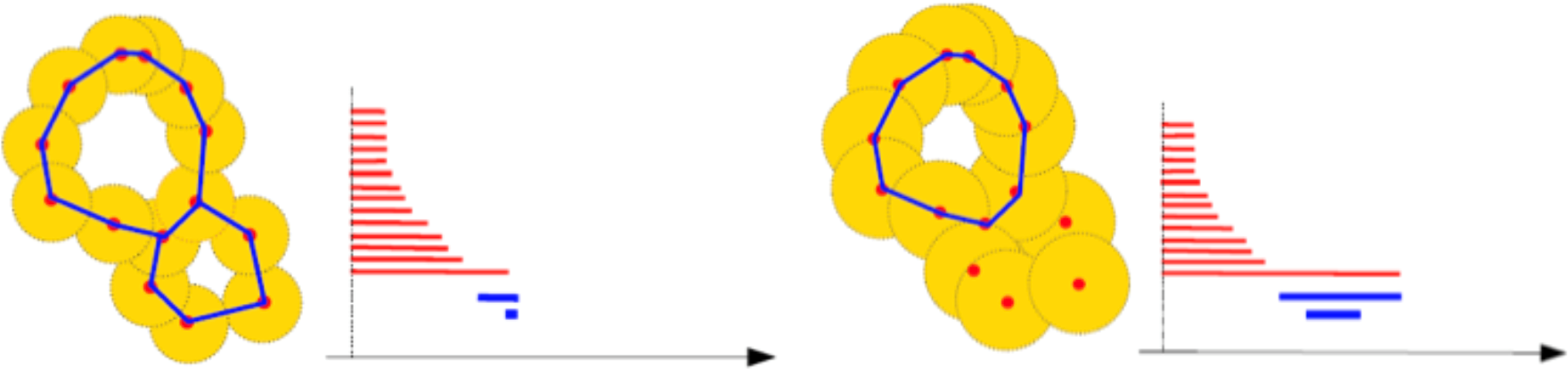} 
\end{tabular}
  \caption{An example of the barcode for $H_{*}(X)$ where $X$ is a set of growing balls. Clearly the sets $\{X_{\epsilon}\}$ form a filtration where $\epsilon$ is the radius of each ball. The rank of $H_k(X_{\epsilon_i})$ equals the numbers of intervals in the barcode for $H_{*}(X)$ intersecting a vertical line $\epsilon = \epsilon_i$ \cite{gudhi:PersistenceRepresentations}. The red bars are zero-order Betti numbers and the blue bars represent the first-order Betti numbers. Different bars are Betti numbers correspond to different generator. Draw a vertical line at $\epsilon_0$ from x-axis and intersect with the bars can find the rank of $H_0(X_{\epsilon_0})$ or $H_1(X_{\epsilon_0})$ for a certain generator.}
  \label{fig:point_betti}
\end{figure*}

Persistent homology is the crucial tool for Topological Data Analysis (TDA), which extracts the high dimensional information from datasets. It provides a general framework to analyze the data regardless of the chosen metric and to compose discrete points into a global structure. TDA is used to analyze and visualize the persistent homology. In the computational process, given a collection of points $\{x_i\}$ in $\mathbb{R}^d$, first, one type of the complex is computed under certain criteria. Then the persistent homology on the complex is determined. As Figure~\ref{fig:point_betti} shows, a persistent barcodes graph is used to visualize the homological structures on the dataset.

A barcodes graph is a graphical representation of $H_k(X)$ as a collection of horizontal line segments in a plane. The horizontal axis corresponds to the radius of the generated complex and the vertical axis is an ordering of the homology generators.  
\subsection{Preliminary Results}
The characterization of homology structure on a neural network and the relationship between Betti numbers and the learning ability of a neural network have been used in some previous work. The work of Bianchini and Scarsellii~\cite{Bianchini2014} gives the upper bounds on the sum of Betti numbers on a binary classification network with a shallow or strictly constrained deep neural network based on Pfaffian sets, with polynomial activation cases only. The bounds are up to $O(\deg(P)^{nl})$, where $n$ is the number of nodes and $l$ is the number of layers, and $\deg(P)$ is the degree of the polynomial. The work of Guss et al~\cite{William2018}. gives an empirical analysis on the effect of the number of hidden units of the first layer on the performance of a classification network using algebraic topology. Other work has focused on finding the Betti number range on various mathematical structures such as semi-algebraic sets (\cite{Basu2001, Gabrielov2004}). Our work extends the mathematical results and applies them to $n$-classifying neural networks. We are able to bound the Betti numbers of each layer, the bound is tight up to $O((l \deg(P))^n)$, and our results are applicable to ReLU. This is archived by extracting the algebraic structure on the layer to form semi-algebraic sets, which is a different approach than using Paffian functions.


\section{Homological Structure of Neural Networks}\label{sec:main}
In this section, we will analyze the homology structure on each layer of a neural network. First, let us define the topological structure on neural architectures. A feedforward $n$-classification neural network $\mathcal{N}: \mathbb{R}^d \rightarrow \{1,2,\cdots, n\}$ is given by the composition 
\begin{equation}\label{eq:neual_network} \mathcal{N}= s \circ f_l \circ \sigma \circ f_{l-1} \cdots \sigma \circ f_1(x)\:.
\end{equation}
Each layer acts as an affine transformation $f_j: \mathbb{R}^{n_j} \to \mathbb{R}^{n_{j+1}}, j= 1,2, \cdots, l - 1$ and $\sigma: \mathbb{R}^{n_{j+1}} \to \mathbb{R}^{n_{j+1}}$ is the activation of this layer. The softmax or sigmoid function on the last layer is denoted as $s:\mathbb{R}^{n} \to \mathbb{R}^n$. Then the network has $l$ layers, the number of hidden units on layer $j$ is $n_j$, and the $l$-th layer is the output layer that maps the input data to $n$ classes. Now consider the output set of the neural network $S = \bigcup_{j=1}^n S(j)$, where $S(j)\subset S$ is the set of the $j$-th class, i.e., $S(j)$ is the image of $s\circ f_l$ in class $j$. Traversing these classified sets back to the $i$-th layer, we obtain the pre-images of $S(j)$ on the $i$-th layer, denoted $S^i(j)$. Then the topological invariant of each layer is characterized by the property of the sets $\{S^i(j)\}_j^i$. However, due to the unknown sign of the elements in the weight matrices, one encounters a system of inequalities when directly analyzing the topological properties of $\{S^i(j)\}_j^i$. To avoid the problem, we have instead focused on the boundary sets $\partial S^i(j)$ of the pre-images instead, which are the solution of a system of equations, and the properties of $\{S^i(j)\}_j^i$ can be characterized by these boundary sets using the Mayer-Vietoris sequence, more details will be presented in the coming sections.\par
The boundary of the set $S^i(j)$ is the ambiguous set associated to the $j$-th class, i.e. the mapping from the $i$-th layer to the output is $F_i= s \circ f_l \circ \sigma \circ f_{l-1} \dots \sigma \circ f_i(x): \mathbb{R}^{n_i} \to \mathbb{R}^n$ for the $n$ classifier and $F_i=[F^1_i, ..., F_i^n]$ has $n$ components. If $x_0$ is mapped to class $j$, then the $j$-th component $F_i^j(x_0)$ is said to have value greater than the other component $F_i^k(x_0), ~k\neq j$. The boundary of the $j$-th class $\partial S^i(j)$ is $\{x \in \mathbb{R}^{n_i} | \exists k \text{ s.t } F_i^k(x) = F_i^j(x), F_i^j(x) \geq F_i^p(x), p\neq j,k\}$. Thus, the set can be decomposed according to the number of equal component functions by a set of equalities and inequalities forming a submanifold. The classical Mayer-Vietoris sequence gives the relation over the homology groups between their intersection and union of two topological manifolds.

\subsection{Algebraic Structure of Dense Layers}
We go further and examine the properties of this boundary set. Every $S^i(j)$ has a decomposition or a covering of semi-algebraic sets. A semi-algebraic set (in the field of real numbers) is a subset $S$ of $\mathbb{R}^n$ defined by a finite sequence of polynomial equations and inequalities, or any finite union of such sets. For simplicity, for the $i$-th dense layer, $f_i(x) = W^i X^{i}+ B^i$, where $W^i$ is the weight matrix and $B^i$ is the bias vector. If the layer is equipped with polynomial activation, $(\sigma\circ f_i)(x) = P(W_i X^{i}+B_i)$, where $P = [Q^i_1, \dots, Q^i_{n_i}]$ is the polynomial activation on the $i$-th layer. Thus, the pre-images of the layers are the composition of the algebraic sets. 
For simplicity, we give the results as two lemmas only; the proof details can be found in the appendix.
\begin{Lemma}\label{lemma:polynomial}
	For a $n$-classification network ($n\geq 2$) has structure defined in~\eqref{eq:neual_network} and $\sigma$ is the \emph{polynomial} activation (degree less than $r$), the boundary set $\partial S^i(j)$ has the following covers
	\begin{equation} \label{eq:polynimial_cover}
	    \partial S^i(j) = \bigcup_{p \neq j}( \{Q^{i}_j - Q^{i}_p = 0\} \cap \{Q^{i}_j - Q^{i}_q \geq 0, q\neq p,j\}) = \bigcup_{p \neq j} S^i_p(j)\:,
	\end{equation}
    where $Q^{i}_j, 1 \leq j \leq n$ is the $j$ component polynomial function on the $i$-th layer with $\deg(Q^{i}_j) \leq r(l-i-1)$. Moreover, for a sub index set $\{\alpha_0,\dots, \alpha_p\}$ of $\{1,2,\dots,j-1,j+1,\dots, n\}$, then $S^i_{\alpha_0\dots\alpha_p}(j) = S^i_{\alpha_0}(j) \cap \dots\cap S^i_{\alpha_p}(j) = \{Q^{i}_j - Q^{i}_{\alpha_k} = 0, k = 0,\dots, p\}\cap \{Q^{i}_j - Q^{i}_q \geq 0, q\neq \alpha_0, \dots, \alpha_p,j\}$.
\end{Lemma}
	
 \begin{Lemma}\label{lemma:relu}
For a $n$-classification network ($n\geq 2$) has structure defined in \eqref{eq:neual_network} with \emph{ReLU} activation whose weight matrices are always full ranks, the pre-images on the boundary of the $j$-th class on the $i$-th layer $\partial S^i(j)$ has the cover
\begin{equation}\label{eq:ReLu}
  \partial S^i(j) = \bigcup_{p \neq j} S^i_p(j),~~\text{and}~~S^i_{\alpha_0\cdots\alpha_p}(j) = S^i_{\alpha_0}(j) \cap \cdots\cap S^i_{\alpha_p}(j)
\end{equation}
for sub index set $\{\alpha_k\}_k$ defined the same as in lemma~\ref{lemma:polynomial}. The cover is determined by the following equations
\begin{align*}
&X^{l-1}_1 = -\widetilde{W}_1^{-1}\widetilde{B}_{\alpha_0\cdots\alpha_p} -\widetilde{W}_1^{-1}\widetilde{W}_2 X^{l-1}_2 \:,\\
&X^{l-2}_1 = -(W^{l-2}_1)^{-1}(\widetilde{B}^{l-1} - B^{l-1}) -(W^{l-2}_1)^{-1}W^{l-2}_2 X^{l-2}_2\:,\\
&\cdots \\
&X^{i}_1 = -(W^{i}_1)^{-1}(\widetilde{B}^{i + 1} - B^{i + 1}) -(W^{i}_1)^{-1}W^{i}_2 X^{i}_2\:,
\end{align*}
where $W^i = (w^i_{ij})\in \mathbb{R}^{n_i\times n_{i-1}}$ is the weight matrix $B^i = [b^i_0,\dots,b^i_{n_i}]$ is the bias vector on the $i$-th layer.
\begin{align*}
&\widetilde{B}_{\alpha_0\dots\alpha_p} =
\left[
b^{l}_{j} - b^{l}_{\alpha_0}, b^{l}_{j} - b^{l}_{\alpha_1},\dots
b^{l}_{j} - b^{l}_{\alpha_p}
\right]^\top\:, \\
&\begin{bmatrix}
\widetilde{W}_1 & \widetilde{W}_2
\end{bmatrix} = 
\widetilde{W}_{\alpha_0\cdots\alpha_p} =
\begin{bmatrix}
w^{l-1}_{j,1} - w^{l-1}_{\alpha_0,1} & w^{l-1}_{j,2} - w^{l-1}_{\alpha_0,2} & \cdots &w^{l-1}_{j,n_{l-1}} - w^{l-1}_{\alpha_0,n_{l-1}}\\
w^{l-1}_{j,1} - w^{l-1}_{\alpha_1,1} & w^{l-1}_{j,2} - w^{l-1}_{\alpha_1,2} & \cdots &w^{l-1}_{j,n_{l-1}} - w^{l-1}_{\alpha_1,n_{l-1}}\\
\vdots       & \vdots        & \ddots& \vdots \\
w^{l-1}_{j,1} - w^{l-1}_{\alpha_p,1} & w^{l-1}_{j,2} - w^{l-1}_{\alpha_p,2} & \cdots &w^{l-1}_{j,n_{l-1}} - w^{l-1}_{\alpha_p,n_{l-1}}
\end{bmatrix} \:,\\
&\widetilde{B}^{i + 1}  =
\begin{bmatrix}
-(W^{i+1}_1)^{-1}(\widetilde{B}^{i+2} - B^{i+2}) \\
0
\end{bmatrix} +
\begin{bmatrix}
-(W^{i+1}_1)^{-1}W^{i+1}_2 \\
I
\end{bmatrix} X^{i+1}_2	= \begin{bmatrix}
X^{i+1}_1 \\
X^{i+1}_2
\end{bmatrix}\:,  
\end{align*}
and $W^k = \left[
{W}^k_1, {W}^k_2
\right]$, $W^{k}_1$ is the reversible sub-matrix of full rank matrix $W^{k}$ and the inequalities.
$$X^{l-1} \in \mathbb{R}_{+}^{n_{l-1}}, X^{l-2} \in \mathbb{R}_{+}^{n_{l-2}}, \cdots,X^{i} \in \mathbb{R}_{+}^{n_{i}} $$
\begin{align*}
\begin{bmatrix}
X^{l}_{j} - X^{l}_{q}
\end{bmatrix} &= 
\begin{bmatrix}
w^{l-1}_{j,1} - w^{l-1}_{q,1} & w^{l-1}_{j,2} - w^{l-1}_{q,2} & \cdots &w^{l-1}_{j,n_{l-1}} - w^{l-1}_{q,n_{l-1}}\\
\end{bmatrix}
\begin{bmatrix}
X^{l-1}_{1}& X^{l-1}_{2}& \dots &X^{l-1}_{n_{l-1}} 
\end{bmatrix}^\top \\ &+ \begin{bmatrix}
b^{l}_{j} - b^{l}_{q}
\end{bmatrix} \geq 0
\end{align*}
where $q \neq \alpha_0, \cdots, \alpha_p, j$.
\end{Lemma}
Using subindex set $\{\alpha_k\}_k$ to build the new cover of $S^i(j)$ is for using generalized Mayer-Vietoris sequence later, such that we can use the Betti numbers of covering sets to bound the Betti numbers of the whole set. 

\subsection{Betti numbers on the Neural Architectures}
Each layer of the neural network was characterized by a semi-algebraic set covering. The work of Basu et al. \cite{Basu2001} derived the upper bound on the Betti numbers for a semi-algebraic set; other work (\cite{Milnor1964,PetOle1949,Thom1965}) derived the upper bounds on the maximum sum of Betti numbers of any algebraic set defined by polynomials. To bound the Betti numbers on each layer, we need to prove the inequalities of the Betti numbers for individual sets and their union. The main steps to establish the relation are: first, $\partial S^i(j) = \overline{S^i(j)} \cap \overline{U/S^i(j)}$, in the polynomial case $U =\mathbb{R}^{n_i}$ and in the ReLU case $U =\mathbb{R}_{+}^{n_i}$. The boundary set can then be described by the closure of pre-images of each class on each layer. Next, we generalize the Mayer-Vietoris sequence to be suitable for $n$ covering a set $\partial S^i(j)$. Second, we use the Mayer-Vietoris sequence to show that the Betti numbers of the set $\partial S^i(j)$ are not larger than the sum of the Betti numbers of the individual cover sets $S^i_{\alpha_o,\dots,\alpha_p}(j)$ in \eqref{eq:polynimial_cover} and \eqref{eq:ReLu}. That is, if $S_1, S_2,..., S_n$ is the cover of the set $S$, then the $k$-th Betti number of the set $b_k(S)$ satisfies.
\begin{equation}\label{eq:betti_sum}
    b_k(S) \leq \sum_{i + j = k}\sum_{J\subset \{1,2,..,n\} \norm{J}=j + 1} b_i(S_J)\:.
\end{equation}
Here $J$ is a subindex set like $\{\alpha_i\}_i$ in Lemma~\ref{lemma:polynomial} and~\ref{lemma:relu}, $\norm{J}$ is the number of sets. This inequality is derived by proving that the rank of the homological group of the set $S$ is less than the sum of the ranks of the intersections of its covers. Finally, we apply the results of Basu et al~\cite{Basu2001}. and Milnor et al~\cite{Milnor1964}. to find upper bounds on Betti numbers. The details of the proofs can be found in the appendix. The main result is derived after all the procedures have been collected.
\begin{Theorem}\label{th:poly}
Suppose a neural network $\mathcal{N}$ has structure as defined in \eqref{eq:neual_network} with \emph{polynomial} activation $P$ ($\deg(P) \leq r$), $L= \{L_i | l \geq i \geq 0, \dim(L_i)=n_i\}$ denotes the layers of the neural network and $n_l=n$, and assume this neural network has a non-increasing structure, i.e., $n_p \leq n_q, p \geq q$, then the Betti numbers of the closure of the $j$-th class pre-image on the $i$-th layer $\overline{S^i(j)}$ are bounded by the following inequalities:
\begin{equation*}
b_k(\overline{S^i(j)})  \\ 
\leq \left\{
\begin{aligned}
&	\sum_{p =0}^{k} \binom{n-1}{p + 1} (3^{n-2-p} - 1) r(l - i - 1) (2r(l - i - 1) - 1)^{n_i - 1},& k < n-2, \\	
&   \left(\sum_{p =0}^{n-3} \binom{n-1}{p + 1} (3^{n-2-p} - 1) + 1\right) r(l - i - 1) (2r(l - i - 1) - 1)^{n_i - 1},&k \geq n-2,
\end{aligned}
\right.
\end{equation*}	
for $i <  l -1$. On the $(l-1)$-th layer
\begin{equation*}
b_k(\overline{S^{l-1}(j)})  \\ 
\leq \left\{
\begin{aligned}
&	\sum_{p =0}^{k} \binom{n-1}{p + 1} (3^{n-2-p} - 1),      & k < n-2, \\	
&   \sum_{p =0}^{n-3} \binom{n-1}{p + 1} (3^{n-2-p} - 1) + 1,&k \geq n-2\:.
\end{aligned}
\right.
\end{equation*}		
\end{Theorem}

\begin{Theorem}\label{th:relu}
Suppose a neural network $\mathcal{N}$ has structure as defined in \eqref{eq:neual_network} with \emph{ReLU} activation, $L= \{L_i | l \geq i \geq 0, \dim(L_i)=n_i\}$ denotes the layers of the neural network and $n_l=n$, and assume this neural network has a non-increasing structure, i.e., $n_p \leq n_q, p \geq q$, then the Betti numbers of the closure of the $j$-th class pre-image on the $i$-th layer $\overline{S^i(j)}$ are bounded by the following inequalities:
\begin{equation*}
b_k(\overline{S^i(j)})  \\ 
\leq \left\{
\begin{aligned}
&	\sum_{p =0}^{k} \binom{n-1}{p + 1} (3^{\sum_{q = l-1}^{i} n_q + n-2-p} - 1)       ,& k < n-2 \:, \\	
&   \sum_{p =0}^{n-2} \binom{n-1}{p + 1} (3^{\sum_{q = l-1}^{i} n_q + n-2-p} - 1) ,& k \geq n-2 \:.
\end{aligned}
\right.
\end{equation*}
\end{Theorem}
\section{Homological Neural Network Architecture Selection}\label{sec:experiment}
In the previous section~\ref{sec:main} we derived the upper bounds of the Betti numbers at each layer. However, the practical use of Betti numbers must apply to the real data. To validate our conclusion and demonstrate the interpretation of the derived results, a dataset MNIST (\cite{lecun2010}) are used for testing. 
The persistent homology is computed using the Python library GUDHI~\cite{gudhi:urm}. We focus on the zero-order and first-order topological features, i.e., $b_0$ and $b_1$ for each class, to characterize the neural network complexity, because the package copes better with lower-dimensional homological features than with higher dimensions.
\subsection{Expression Ability}
\begin{figure*}[!ht]
  \centering
  \setlength\tabcolsep{2pt} 
  \def\arraystretch{1} 
  \newcommand{\mywidth}{0.3\textwidth} 
  \newcommand{\mywidthlr}{0.32\textwidth} 
  \newcommand{\mywidths}{0.10\textwidth}
  \newcolumntype{M}{>{\centering\arraybackslash}m{\mywidth}}
  \newcolumntype{Y}{>{\centering\arraybackslash}m{\mywidthlr}}
\begin{tabular}{MYY}
\multirow{3}{*}[9.5ex]{\includegraphics[width = \mywidth]{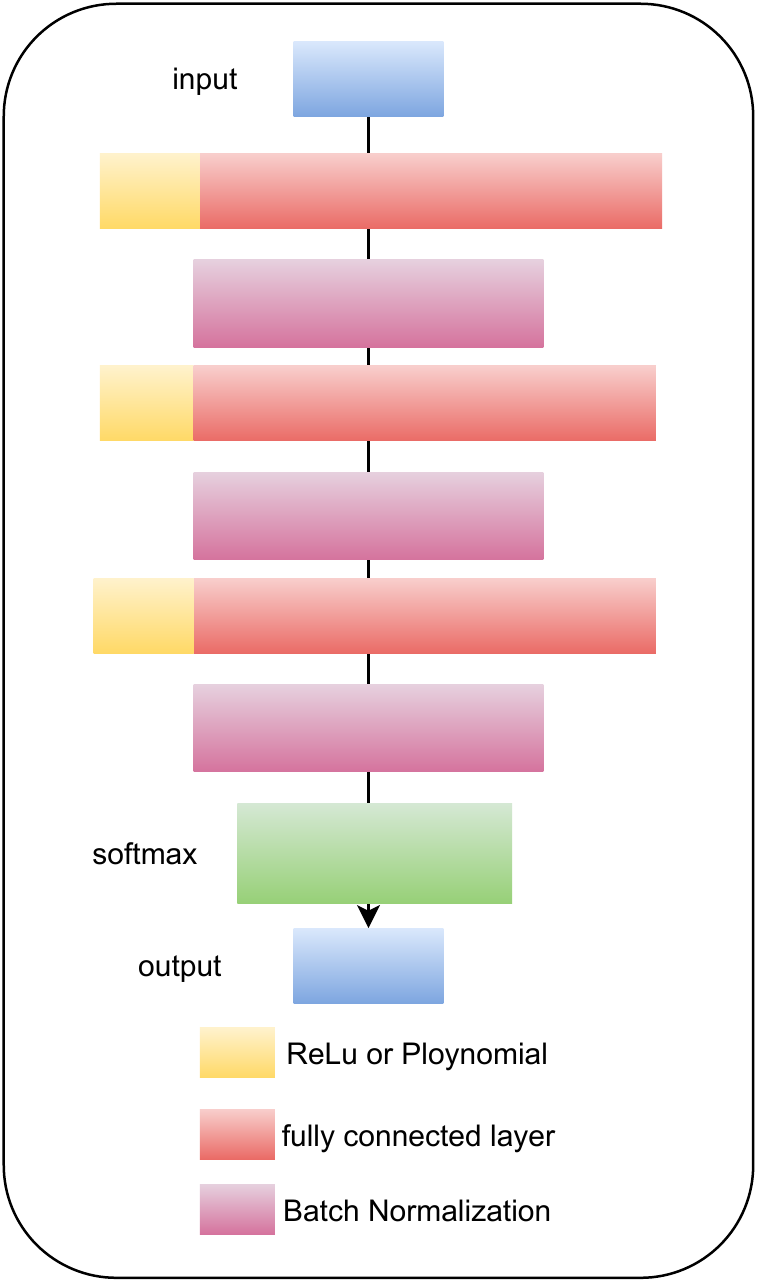}} & 
\includegraphics[width = \mywidth]{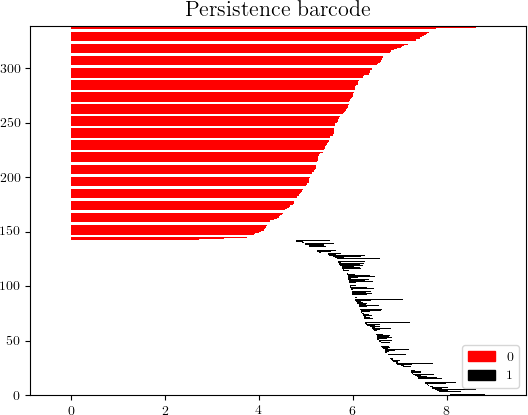} &
\includegraphics[width = \mywidth]{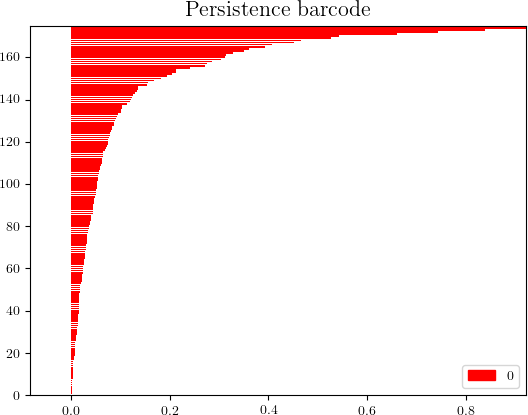} \\
& (b) input data Betti numbers & (c) inefficient presentation \\ 
& \includegraphics[width = \mywidth]{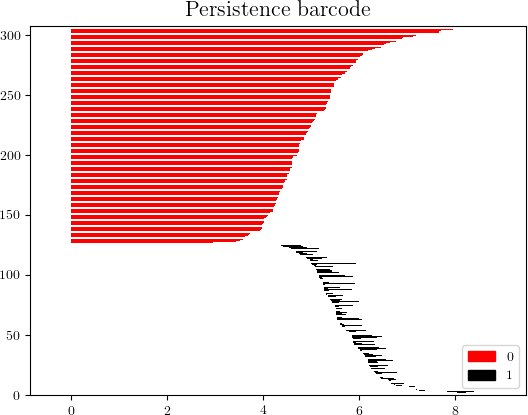} &
\includegraphics[width = \mywidthlr]{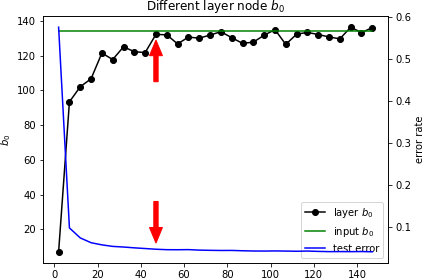} \\
(a) & (d) efficient representation & (e) output $b_0$ and accuracy
\end{tabular}
  \caption{((a) Network structure: the network has 3 fully connected layers with ReLU activation and batch normalization. (b)-(d) A network with sufficient expression ability should have sufficient topology complexity to represent the geometry of the input data. (e) The varying number of hidden units affects the complexity of the network. If a network can achieve a certain accuracy, the layer output (the pre-images of the classes) should have $b_0$ converging with the $b_0$ of the input data.}
  \label{fig:relu}
\end{figure*}

We validate our conjecture that the Betti numbers of the layers reflect the expressiveness of the network. We build networks with the architecture shown in Figure~\ref{fig:relu} with different hidden units; each dense layer has the same hidden units in each network. The layer number $l = 4$ and we tested the hidden units $n_i$ for $i=1,2,3$ from 2 to 147. We take the output of the ten different classes from the third layer (the layer before Softmax) and computed their zero-order and first-order Betti numbers $b_0^i$, $b_1^i$ for $i = 1,\dots,10$. Note that batch normalization is added because the persistent homology barcodes are computed with respect to a certain radius.

The $b_0$ of class 4 and the persistent homology barcodes are shown in Figure~\ref{fig:relu}. In the inefficient representation, the persistent barcode graph behaves dissimilarly and cannot interpret the first-order features, also shown in Figure~\ref{fig:inefficient}. The efficient representation occurs when the layers equip enough hidden units and can express more topological structure than the input data. Our presented theorems prove that expressiveness is determined by the number of layers and the number of hidden units of a dense neural network. Note that the dense layers are affine transformations and should not increase the topological structures of the input data. 
Thus, when the node number increases, the Betti numbers of the layer output will not unlimitedly increase, they should be close to the Betti numbers of the input data. We take the maximum $b_0$ with a minimum radius of the 10 classes pre-images from the third layer and compare them with the input data $b_0$, plotted in Figure~\ref{fig:relu} (c). If the network achieves a certain accuracy, the Betti numbers are close to the input data.
\begin{figure*}[!ht]
  \centering
  \setlength\tabcolsep{2pt} 
  \def\arraystretch{0} 
  \newcommand{\mywidth}{0.25\textwidth} 
  \newcommand{\mywidthlr}{0.43\textwidth} 
  \newcommand{\mywidths}{0.45\textwidth}
  \newcolumntype{H}{>{\centering\arraybackslash}m{\mywidth}}
  \newcolumntype{Y}{>{\centering\arraybackslash}m{\mywidths}}
\begin{tabular}{YY}
\includegraphics[width = \mywidths]{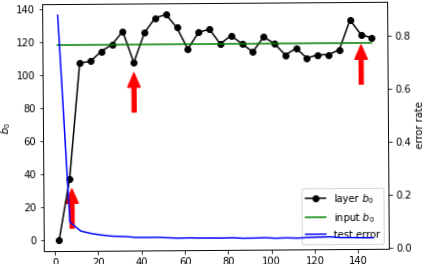} &
\includegraphics[width = \mywidthlr]{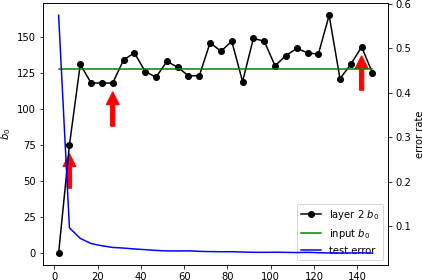} \\
output $b_0$ class 6 & output $b_0$ class 9
\end{tabular}
  \caption{ Investigate the output $b_0$ of class 6 and class 9 on the third layer of the neural network with different layer size. The $b_0$ increases with layer size and tries to be consistent with the input data $b_0$.}
  \label{fig:outputb0}
\begin{tabular}{HHHH}
\includegraphics[width = \mywidth]{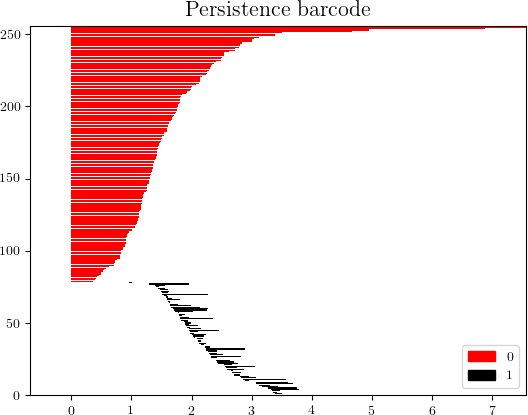}&
\includegraphics[width = \mywidth]{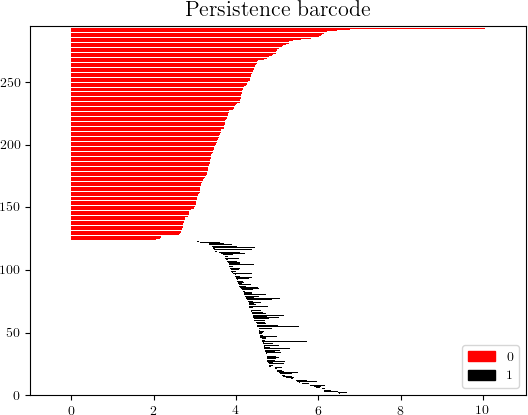}&
\includegraphics[width = \mywidth]{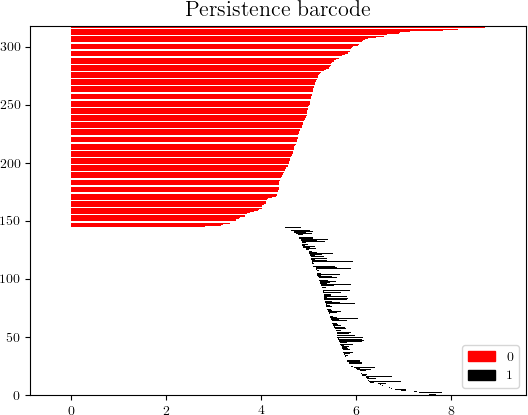} & 
\includegraphics[width= \mywidth]{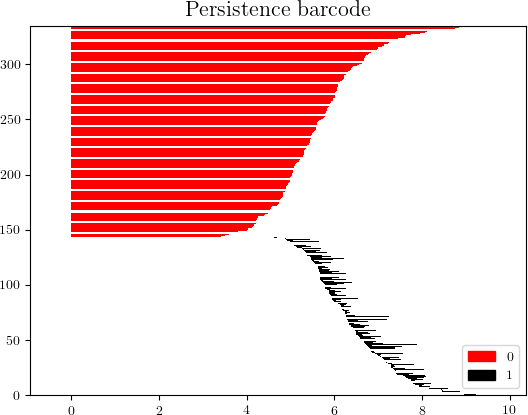}\\
 node = 7 & node = 37 & node = 142 & input class 6 \\
\includegraphics[width = \mywidth]{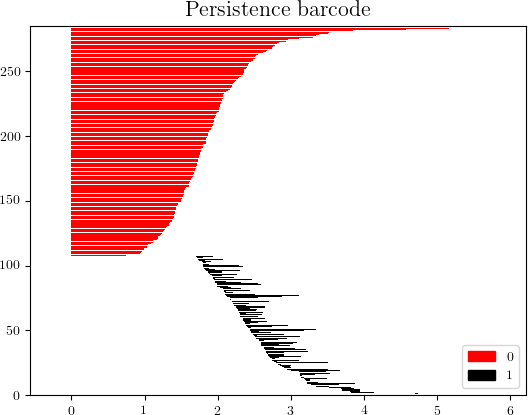}&
\includegraphics[width = \mywidth]{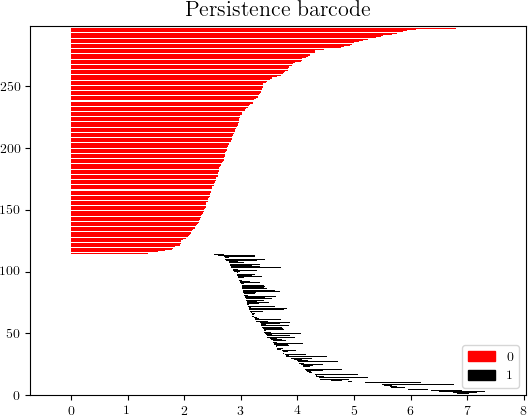}&
\includegraphics[width = \mywidth]{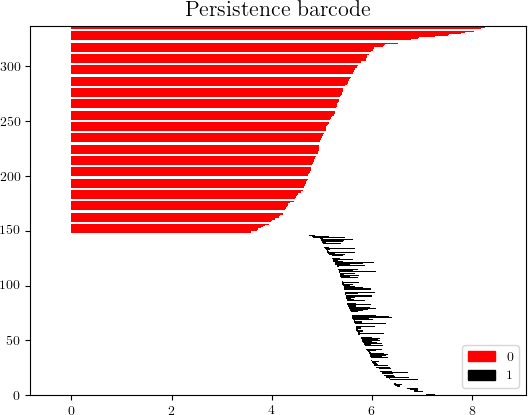} & 
\includegraphics[width= \mywidth]{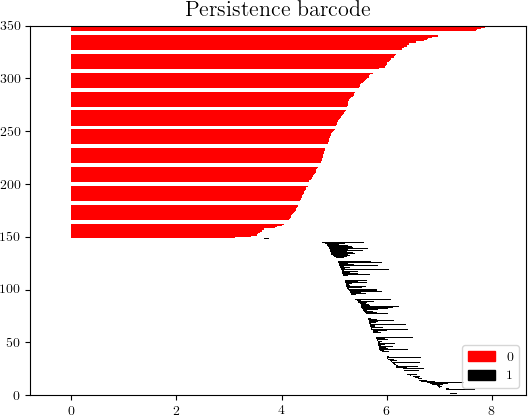}\\
  node = 7 & node = 27 & node = 142 & input class 9 
\end{tabular}
  \caption{The graphs in the first row are persistent barcodes of class 6 and the second row is of class 9. The graphs in the first three columns correspond to the node numbers pointed to by the three red arrows in each graph in Figure~\ref{fig:outputb0}. The graphs in the last column are the barcodes of the input classes. The inefficient representation occurs when $b_0$ is significantly smaller than the input data $b_0$ and the persistent barcode graph is different from the barcode graph of the input data barcodes graph. When the layer interprets the input data efficiently, the output and input barcode graphs have the similar structure.}
  \label{fig:layer3class69}
  
\end{figure*}

The Figure~\ref{fig:outputb0} shows the $b_0$ in different layer size with ReLU activation of class 6 and class 9. As accuracy increases, the $b_0$ tents converge to the input data $b_0$. And the Figure~\ref{fig:layer3class69} shows the persistent homology barcodes of certain layer sizes. 
\begin{figure*}[!ht]
  \centering
  \setlength\tabcolsep{2pt} 
  \def\arraystretch{0} 
  \newcommand{\mywidth}{0.44\textwidth} 
  \newcommand{\mywidthlr}{0.25\textwidth} 
  \newcommand{\mywidths}{0.23\textwidth}
  \newcolumntype{K}{>{\centering\arraybackslash}m{\mywidthlr}}
  \newcolumntype{Y}{>{\centering\arraybackslash}m{\mywidths}}
\begin{tabular}{KYYY}
\includegraphics[width = \mywidthlr]{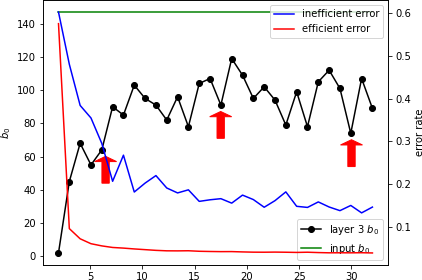} &
\includegraphics[width = \mywidths]{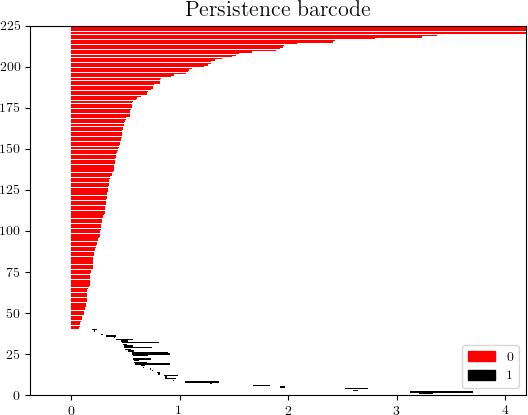}&
\includegraphics[width = \mywidths]{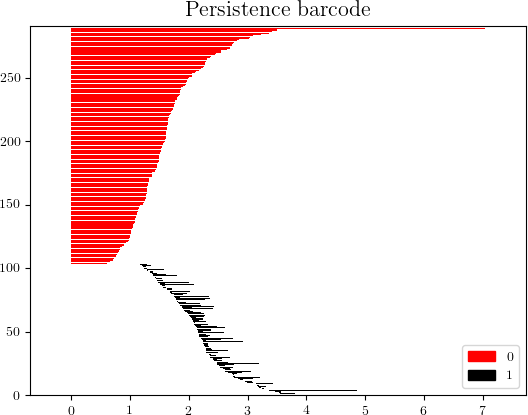}&
\includegraphics[width = \mywidths]{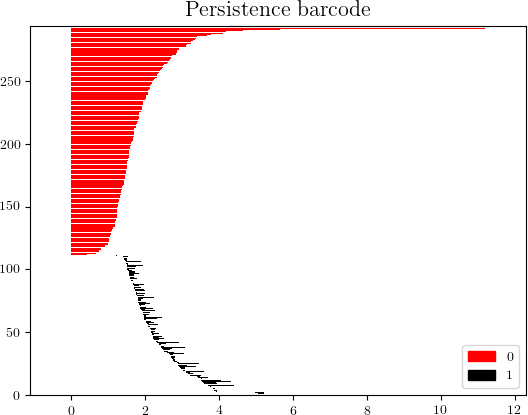} \\
output $b_0$  & node = 5 & node = 7 & node = 17 
\end{tabular}
  \caption{When a neural network are not well trained, the layer fails to represent the homology structure of the input data. Take class 9 as an example, the input data barcodes graph is shown in Figure~\ref{fig:layer3class69}. The blue line is the test error of the inefficient representation and the red line is the test error of a well trained network. Plot the barcodes graphs at the three red arrows, they all show different structures from inpout data.}
  \label{fig:inefficient}
\end{figure*}

\subsection{Betti Numbers Bounding Interpretation}
Theorem~\ref{th:poly} and Theorem~\ref{th:relu} show that the upper bounds of Betti numbers for each layer decrease as the layer index increases, i.e., if all layers are of equal size, then the upper bounds of Betti numbers are descending along with the layers. Note that if there is no other operation between two dense layers that can increase the topological structure of the input, e.g., only fully connected layers in the network, since the dense layer only applies an affine transformation to the data, it will not increase the topological feature. This means that a neural network should be equipped with enough hidden units of the first dense layer to have sufficient topological complexity to represent the input data. As a decreasing networks, even though without any activation, the affine transforamtion is a projection from high dimensional space to a low dimensional space, the betti numbers will naturally decreasing. The same conclusion was drawn by Guss et al. in~\cite{William2018}, who proposed that the choice of the size of the first layer determines the learning ability of the network. Our theoretical work is consistent with their experimental results.\par
\begin{figure*}[!ht]
  \centering
  \setlength\tabcolsep{2pt} 
  \def\arraystretch{1} 
  \newcommand{\mywidth}{0.48\textwidth} 
  \newcommand{\mywidthlr}{0.25\textwidth} 
  \newcommand{\mywidths}{0.2\textwidth}
  \newcolumntype{C}{>{\centering\arraybackslash}m{\mywidth}}
  \newcolumntype{Y}{>{\centering\arraybackslash}m{\mywidths}}
\begin{tabular}{CC}
\includegraphics[width = \mywidth]{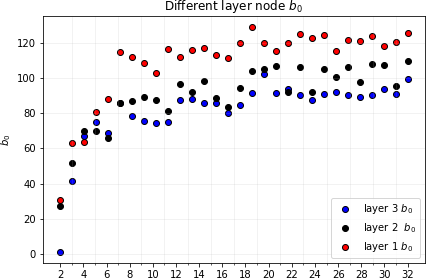} &
\includegraphics[width = \mywidth]{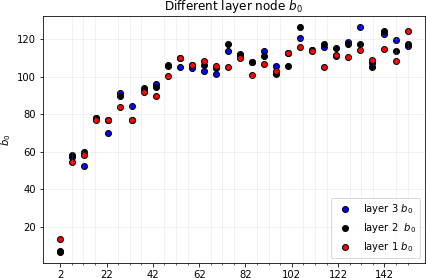} \\
\centering (a) & \centering (b)
\end{tabular}
  \caption{(a) For a network with three equal-size dense layers (with polynomial activation), change the layer size, Betti numbers of the different layers are descending along the layers. (b) The homology structures within the layers are preserved if the number of hidden nodes is sufficient and the complexity of the input exceeds the expressiveness of a network with fewer nodes. Here are the layers with ReLu activation.}
  \label{fig:layer_decrease}
\end{figure*}

We validate our interpretation against the MNIST dataset, as Figure~\ref{fig:layer_decrease} shows. The average $b_0$ with minimum radius from the 10 classes on each layer are taken. In the left graph, relatively small nodes are taken so that the expression capacities of the layers (with polynomial activation) may not exceed the input data. The output Betti numbers on each layer tend to decrease along with the layers. In the case of  ReLU activation, we see how Betti numbers are kept among layers. As larger layer sizes are chosen, the complexity of the network exceeds the complexity of the input data, hence the $b_0$ are quite similar among layers since they are bounded by the input data $b_0$.
\par
\begin{figure*}[!ht]
  \centering
  \setlength\tabcolsep{2pt} 
  \def\arraystretch{0} 
  \newcommand{\mywidth}{0.48\textwidth} 
  \newcommand{\mywidthlr}{0.25\textwidth} 
  \newcommand{\mywidths}{0.2\textwidth}
  \newcolumntype{L}{>{\centering\arraybackslash}m{\mywidth}}
  \newcolumntype{Y}{>{\centering\arraybackslash}m{\mywidths}}
\begin{tabular}{LL}
\includegraphics[width = \mywidth]{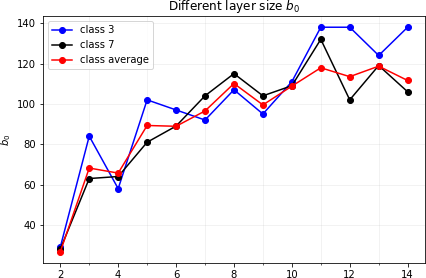} &
\includegraphics[width = \mywidth]{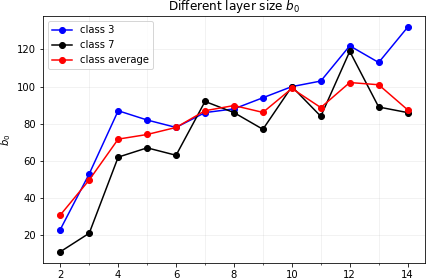} \\
\centering (a) & \centering (b)
\end{tabular}
  \caption{Given a three dense layers network, examine the $b_0$ on one of the three layers by changing the layer size. The Betti numbers of the different layer sizes are in ascending order. The left graph is the layer with polynomial activation and the right graph is the layer with ReLU activation.}
  \label{fig:increase_node}
\end{figure*}

The second observation from the derived theorems is that the upper bounds increase with the size of the layer. In other words, layers with more hidden units are more likely to have the efficient expressive ability. This observation has been known for a long time. Our results provide a mathematical explanation for the conjecture. In Figure~\ref{fig:increase_node}, different hidden units were chosen for a layer and the $b_0$ on that layer was calculated. Note that relatively fewer node numbers were chosen to ensure that the homology complexity of the layer does not exceed the complexity of the input data. Otherwise, as explained earlier, the layer, although having more complexity, would only maximally represent the complexity of the input data.


\section{Conclusion and Future Work}\label{sec:conclusion}
This work follows from the seminal work (\cite{Cybenko1989, Eldan2015, Bianchini2014}) that attempted to understand the expressiveness of neural network architectures and to exploit the expressive capacity of each layer of a neural network in the language of topological complexity. Our algebraic view characterizes the expressivity through topological invariants: Betti numbers and we derive the upper bounds on the Betti numbers on each layer of a neural network. The empirical question of the performance and learnability of a network can then be explained mathematically. Compared to previous work, our results provide explicit upper bounds on the topological complexity at each layer of a network and provide theoretical evidence for observed conjectures in the choice of a neural network's architecture that goes beyond empirical conclusions. The presented work extracts the data and neural networks topological features and assigns a computable measure to describe the architecture power. Then the architecture selection problem is boiled down to chose a suitable network size according to the measure. Moreover, our results and conjectures are verified in a real-world dataset. \par
There are several possible avenues for the following future research. First, it is possible to extend the holomogical complexity analysis to other types of layers such as convolutional layers, since convolutional layers share some linearities with dense layers~\cite{Ma2017}. Second, the bounds on Betti numbers could be further tightened. However, further tightening of the bounds depends on a better related spectral theory that can bound the Betti numbers on semi-algebraic sets, and a smaller covering of the pre-images can be found. Third, the analysis of lower bounds is also a promising direction to study the expressivity of neural networks.

\nocite{*}
\bibliographystyle{plain}
\bibliography{references}
\newpage

\appendix
\section{Appendix}
\subsection{Proof of Lemma~\ref{lemma:polynomial}}
\begin{Lemma*} 
	For a $n$-classification network ($n\geq 2$) has structure defined in~\eqref{eq:neual_network} and $\sigma$ is the \emph{polynomial} activation (degree less than $r$), the boundary set $\partial S^i(j)$ has the following covers
		$$\partial S^i(j) = \bigcup_{p \neq j}(\{Q^{i}_j - Q^{i}_p = 0\}\cap \{Q^{i}_j - Q^{i}_q \geq 0, q\neq p,j\}) = \bigcup_{p \neq j} S^i_p(j)\:,$$
		and moreover
		\begin{equation*}
		\begin{aligned}
		S^i_{\alpha_0\cdots\alpha_p}(j) &= S^i_{\alpha_0}(j) \cap \cdots\cap S^i_{\alpha_p}(j) \\
		&= \{(Q^{i}_j - Q^{i}_{\alpha_0})= 0, \cdots, Q^{i}_j - Q^{i}_{\alpha_p}) = 0\} \cap \{Q^{i}_j - Q^{i}_q \geq 0, q\neq \alpha_0, \cdots, \alpha_p,j\} \:,
		\end{aligned}
		\end{equation*}
		where $Q^{i}_j, 1 \leq j \leq n$ is the $j$ component polynomial function on the $i$-th layer with $\deg(Q^{i}_j) \leq r(l-i-1)$.
	\end{Lemma*}
	\begin{proof}
Consider the output before the sigmoid or softmax activation, for $W^i$ is the weight matrix on layer $i$ and $B^i$ is the translation,
		\begin{equation*}
		\begin{aligned}
			X^{l} & =  (W^{l-1} X^{l-1} + B^{l- 1})) \\
			      & =  (W^{l-1} \left(P\left(W^{l-2} X^{l-2} + B^{l- 2}\right)\right)) \\
			      & =  (W^{l-1} (P(W^{l-2} P(W^{l-3} X^{l-3} + B^{l- 3}) + B^{l- 2}))) \\
			      & = \cdots \\
			      & =  (W^{l-1} (P(W^{l-2} P(\cdots P(W^{i} X^{i} + B^{i}) + \cdots + B^{l- 3}) + B^{l- 2}))) \:,
		\end{aligned}
		\end{equation*}
	where $P$ is a polynomial activation and $deg(P) \leq r$, the last formula is endowed with $l - i -1$ polynomial composition, then $\deg(X^{l}(X^{i})) \leq r(l-i-1)$ and these polynomials are denoted by $Q^{i} = \{Q^{i}_1,...,Q^{i}_n\}$. The $j$-th boundary on the $i$-th layer has the cover
	$$\partial S^i(j) = \bigcup_{p \neq j}(\{(Q^{i}_j - Q^{i}_p) = 0\}\cap \{Q^{i}_j - Q^{i}_q \geq 0, q\neq p,j\}) = \bigcup_{p \neq j} S^i_p(j)$$
	and it is clear that
	\begin{equation*}
	\begin{aligned}
		S^i_{\alpha_0\cdots\alpha_p}(j)&= S^i_{\alpha_0}(j) \cap \cdots\cap S^i_{\alpha_p}(j) \\
		&= \{(Q^{i}_j - Q^{i}_{\alpha_0}) = 0, \cdots, (Q^{i}_j - Q^{i}_{\alpha_p}) = 0\}\cap \{Q^{i}_j - Q^{i}_q \geq 0, q\neq \alpha_0, \cdots, \alpha_p,j\}
	\end{aligned}
	\end{equation*}
	is a semi-algebraic set.
	\end{proof}
	
\subsection{Proof of Lemma~\ref{lemma:relu}}
 	\begin{Lemma*}
		For a $n$-classification network ($n\geq 2$) has structure defined in \eqref{eq:neual_network} with \emph{ReLU} activation whose weight matrices are always full ranks, the pre-images on the boundary of the $j$-th class on the $i$-th layer $\partial S^i(j)$ has the cover
		$$\partial S^i(j) = \bigcup_{p \neq j} S^i_p(j)\:,$$
		and furthermore
		$$S^i_{\alpha_0\cdots\alpha_p}(j) = S^i_{\alpha_0}(j) \cap \cdots\cap S^i_{\alpha_p}(j)\:,$$ 
		which is determined by the following equations 
	\begin{equation*}
	\begin{aligned}
	X^{l-1}_1 &= -\widetilde{W}_1^{-1}\widetilde{B}_{\alpha_0\cdots\alpha_p} -\widetilde{W}_1^{-1}\widetilde{W}_2 X^{l-1}_2 \\
	X^{l-2}_1 &= -(W^{l-2}_1)^{-1}(\widetilde{B}^{l-1} - B^{l-1}) -(W^{l-2}_1)^{-1}W^{l-2}_2 X^{l-2}_2\\
	\cdots& \\
	X^{i}_1 &= -(W^{i}_1)^{-1}(\widetilde{B}^{i + 1} - B^{i + 1}) -(W^{i}_1)^{-1}W^{i}_2 X^{i}_2 \:,
	\end{aligned}
	\end{equation*} 
	where
	\begin{equation*}
	\widetilde{B}_{\alpha_0\cdots\alpha_p} =
	\left[ \begin{array}{c} 
	b^{l}_{j} - b^{l}_{\alpha_0} \\
	b^{l}_{j} - b^{l}_{\alpha_1}\\
	\vdots\\
	b^{l}_{j} - b^{l}_{\alpha_p}
	\end{array}
	\right]
	\end{equation*} 	
	\begin{equation*}
	\left[ \begin{array}{cc}
	\widetilde{W}_1 & \widetilde{W}_2
	\end{array}
	\right] = 
	\widetilde{W}_{\alpha_0\cdots\alpha_p} =
	\left[ \begin{array}{cccc}
	w^{l-1}_{j,1} - w^{l-1}_{\alpha_0,1} & w^{l-1}_{j,2} - w^{l-1}_{\alpha_0,2} & \cdots &w^{l-1}_{j,n_{l-1}} - w^{l-1}_{\alpha_0,n_{l-1}}\\
	w^{l-1}_{j,1} - w^{l-1}_{\alpha_1,1} & w^{l-1}_{j,2} - w^{l-1}_{\alpha_1,2} & \cdots &w^{l-1}_{j,n_{l-1}} - w^{l-1}_{\alpha_1,n_{l-1}}\\
	\vdots       & \vdots        & \ddots& \vdots \\
	w^{l-1}_{j,1} - w^{l-1}_{\alpha_p,1} & w^{l-1}_{j,2} - w^{l-1}_{\alpha_p,2} & \cdots &w^{l-1}_{j,n_{l-1}} - w^{l-1}_{\alpha_p,n_{l-1}}
	\end{array} 
	\right ] \\
	\end{equation*}
		\begin{equation*}
	\widetilde{B}^{i + 1}  =
	\left[ \begin{array}{c}
	-(W^{i+1}_1)^{-1}(\widetilde{B}^{i+2} - B^{i+2}) \\
	0
	\end{array}
	\right] +
	\left[ \begin{array}{c}
	-(W^{i+1}_1)^{-1}W^{i+1}_2 \\
	E
	\end{array}
	\right] X^{i+1}_2	= \left[ \begin{array}{c}
	X^{i+1}_1 \\
	X^{i+1}_2
	\end{array}
	\right]	  
	\end{equation*}
	
	\begin{equation*}
	{W}^{k} = \left[ \begin{array}{cc}
	{W}^k_1 & {W}^k_2
	\end{array}
	\right] \:.
	\end{equation*}
	$W^{k}_1$ is the reversible sub-matrix of full rank matrix $W^{k}$ and the inequalities.
	$$X^{l-1} \in \mathbb{R}_{+}^{n_{l-1}}, X^{l-2} \in \mathbb{R}_{+}^{n_{l-2}}, \cdots,X^{i} \in \mathbb{R}_{+}^{n_{i}} $$
	\begin{equation*}
	\begin{aligned}
	\left[ \begin{array}{c} 
	X^{l}_{j} - X^{l}_{q}  \\ 
	\end{array}
	\right]
	&= 
	\left[ \begin{array}{cccc}
	w^{l-1}_{j,1} - w^{l-1}_{q,1} & w^{l-1}_{j,2} - w^{l-1}_{q,2} & \cdots &w^{l-1}_{j,n_{l-1}} - w^{l-1}_{q,n_{l-1}}\\
	\end{array}
	\right ] \\
	&\times 
	\left[ \begin{array}{c} 
	X^{l-1}_{1} \\
	X^{l-1}_{2} \\
	\vdots\\
	X^{l-1}_{n_{l-1}} 
	\end{array} 
	\right] 
	+
	\left[ \begin{array}{c} 
	b^{l}_{j} - b^{l}_{q} \\
	\end{array}
	\right] \\
	& \geq 0 \:,
	\end{aligned}
	\end{equation*}
	where $q \neq \alpha_0, \cdots, \alpha_p, j$.
	\end{Lemma*}
\begin{proof}
	Calculate the boundary determined by the linear equations from the output and represent the pre-images layer by layer. The boundary of the output is determined by the following linear equations, regardless of which activation function is used, sigmoid or softmax function. Assume that the linear equations are solvable.
		\begin{equation*}
		\begin{aligned}
		\left[ \begin{array}{c} 
		X^{l}_{j} - X^{l}_{\alpha_0}  \\
		X^{l}_{j} - X^{l}_{\alpha_1} \\
		\vdots\\
		X^{l}_{j} - X^{l}_{\alpha_p} 
		\end{array}
		\right]
		 &= 
			\left[ \begin{array}{cccc}
			w^{l-1}_{j,1} - w^{l-1}_{\alpha_0,1} & w^{l-1}_{j,2} - w^{l-1}_{\alpha_0,2} & \cdots &w^{l-1}_{j,n_{l-1}} - w^{l-1}_{\alpha_0,n_{l-1}}\\
			w^{l-1}_{j,1} - w^{l-1}_{\alpha_1,1} & w^{l-1}_{j,2} - w^{l-1}_{\alpha_1,2} & \cdots &w^{l-1}_{j,n_{l-1}} - w^{l-1}_{\alpha_1,n_{l-1}}\\
			\vdots       & \vdots        & \ddots& \vdots \\
			w^{l-1}_{j,1} - w^{l-1}_{\alpha_p,1} & w^{l-1}_{j,2} - w^{l-1}_{\alpha_p,2} & \cdots &w^{l-1}_{j,n_{l-1}} - w^{l-1}_{\alpha_p,n_{l-1}}
			\end{array} 
			\right ] \\
			&\times 
		\left[ \begin{array}{c} 
		X^{l-1}_{1} \\
		X^{l-1}_{2} \\
		\vdots\\
		X^{l-1}_{n_{l-1}} 
		\end{array} 
		\right] 
		 +
		\left[ \begin{array}{c} 
		b^{l}_{j} - b^{l}_{\alpha_0} \\
		b^{l}_{j} - b^{l}_{\alpha_1}\\
		\vdots\\
		b^{l}_{j} - b^{l}_{\alpha_p}
		\end{array}
		\right] \\
		&=0\:.
		\end{aligned}
		\end{equation*}
		The inequalities
		\begin{equation*}
		\begin{aligned}
		\left[ \begin{array}{c} 
		X^{l}_{j} - X^{l}_{q}  \\ 
		\end{array}
		\right]
		&= 
		\left[ \begin{array}{cccc}
		w^{l-1}_{j,1} - w^{l-1}_{q,1} & w^{l-1}_{j,2} - w^{l-1}_{q,2} & \cdots &w^{l-1}_{j,n_{l-1}} - w^{l-1}_{q,n_{l-1}}\\
		\end{array}
		\right ] \\
		&\times 
		\left[ \begin{array}{c} 
		X^{l-1}_{1} \\
		X^{l-1}_{2} \\
		\vdots\\
		X^{l-1}_{n_{l-1}} 
		\end{array} 
		\right] 
		+
		\left[ \begin{array}{c} 
		b^{l}_{j} - b^{l}_{q} \\
		\end{array}
		\right] \\
		& \geq 0\:,
		\end{aligned}
		\end{equation*}
		where $q \neq \alpha_0, \cdots, \alpha_p, j$.
		
		Denote the above equalities and inequalities
		$$\widetilde{W}_{\alpha_0\cdots\alpha_p} X^{l-1} + \widetilde{B}_{\alpha_0\cdots\alpha_p} = 0$$
		$$\widetilde{W}_{\hat{\alpha}_0\cdots\hat{\alpha}_p} X^{l-1} + \widetilde{B}_{\hat{\alpha}_0\cdots\hat{\alpha}_p} \geq 0$$
		where $\hat{\alpha}$ means the omission of index $\alpha$.
		Without loss of the generality, let the first $p+1$-block of the square sub-matrix be reversible,
		\begin{equation*}
		  \left[ \begin{array}{cc}
		   \widetilde{W}_1 & \widetilde{W}_2
		   \end{array}
		  \right] 
		  \left[ \begin{array}{c}
		  X^{l-1}_1 \\
	      X^{l-1}_2
	      \end{array}
		  \right] 
		  = \widetilde{W}_1 X^{l-1}_1 + \widetilde{W}_2 X^{l-1}_2 = -\widetilde{B}_{\alpha_0\cdots\alpha_p}	\:.	  
		\end{equation*}
		Then the solution has the form
		\begin{equation*}
		\left[ \begin{array}{c}
		X^{l-1}_1 \\
		X^{l-1}_2
		\end{array}
		\right] =
		\left[ \begin{array}{c}
		-\widetilde{W}_1^{-1}\widetilde{B}_{\alpha_0\cdots\alpha_p} \\
		0
		\end{array}
		\right] +
		\left[ \begin{array}{c}
		-\widetilde{W}_1^{-1}\widetilde{W}_2 \\
		E
		\end{array}
		\right] X^{l-1}_2 \:.		  
		\end{equation*}
		Denote the above vector by $\widetilde{B}^{l-1}$.
		$$\{X^{l-1} \in \mathbb{R}^{n_{l-1}} | X^{l-1}_1 = -\widetilde{W}_1^{-1}\widetilde{B}_{\alpha_0\cdots\alpha_p} -\widetilde{W}_1^{-1}\widetilde{W}_2 X^{l-1}_2\} \bigcap \mathbb{R}_{+}^{n_{l-1}}\:, $$
		and inequalities
		$$\widetilde{W}_{\hat{\alpha}_0\cdots\hat{\alpha}_p} X^{l-1} + \widetilde{B}_{\hat{\alpha}_0\cdots\hat{\alpha}_p} \geq 0 \:.$$
		This is obviously a semi-algebraic set. For $X^{l-1}$ going through the ReLU activation, the pre-images of the boundary in the $(l-1)$-th layer which is mapped from the $(l-2)$-th layer have the following form.
		\begin{equation*}
		\left[ \begin{array}{c}
		X^{l-2}_1 \\
		X^{l-2}_2
		\end{array}
		\right] =
		\left[ \begin{array}{c}
		-(W^{l-2}_1)^{-1}(\widetilde{B}^{l-1} - B^{l-1}) \\
		0
		\end{array}
		\right] +
		\left[ \begin{array}{c}
		-(W^{l-2}_1)^{-1}W^{l-2}_2 \\
		E
		\end{array}
		\right] X^{l-2}_2\:.		  
		\end{equation*}
		The following semi-algebraic set is obtain
		\begin{equation*}
		\begin{aligned}
		S^i_{\alpha_0\cdots\alpha_p}(j) = 
		&\{X^{l-1} \in \mathbb{R}_{+}^{n_{l-1}}, X^{l-2} \in \mathbb{R}_{+}^{n_{l-2}} | \\
		X^{l-1}_1 &= -\widetilde{W}_1^{-1}\widetilde{B}_{\alpha_0\cdots\alpha_p} -\widetilde{W}_1^{-1}\widetilde{W}_2 X^{l-1}_2,\\
		X^{l-2}_1 &= -(W^{l-2}_1)^{-1}(\widetilde{B}^{l-1} - B^{l-1}) -(W^{l-2}_1)^{-1}W^{l-2}_2 X^{l-2}_2 \\
		0 &\leq \widetilde{W}_{\hat{\alpha}_0\cdots\hat{\alpha}_p} X^{l-1} + \widetilde{B}_{\hat{\alpha}_0\cdots\hat{\alpha}_p}
		\} \:.
		\end{aligned}
		\end{equation*} 
	    Then the results can be derived by induction.
\end{proof}

\subsection{Related Results}
In this section we list the Betti number bounding on the semi-albegraic set from previous work.
 	\begin{Lemma}[\cite{Basu2001}]\label{lemma:Basu}
 		Let $\mathbf{R}$ be a real closed field and let $V \subset \mathbf{R}^k$ be the set defined by the conjunction of of $n$ inequalities $\mathbf{P}_1 \geq 0, \cdots, \mathbf{P}_n \geq 0$,
 		$\mathbf{P}\in \mathbf{R}[X_1,...,X_k], deg(\mathbf{P}_i) \leq d, 1\leq i\leq n$, contained in a variety $\mathbf{Z(Q)} = \{\bm{X} \in \mathbb{R}^{k^j}|\mathbf{Q}(\bm{X}) = 0\}$, where $\mathbf{Q}$ is a polynomial of real dimension $k'$ with $deg(\mathbf{Q}) \leq d$. Then, for all $i, 0 \leq i \leq k'$
 		$$b_i(V) \leq \sum_{j=1}^{n} \binom{n}{j} 2^{j} d(2d - 1)^{k- 1}= (3^n - 1)d(2d - 1)^{k- 1}\:.$$	
 	\end{Lemma}
 	
 \begin{Lemma}[\cite{Milnor1964}, \cite{PetOle1949}, \cite{Thom1965}]\label{lemma:Oleinik}
 		 Let $B(k,d)$ be the maximum of the sum of Betti numbers of any algebraic set defined by polynomials of degree $d$ in $\mathbf{R}^k$, then
		$$B(k,d) \leq d(2d-1)^{k-1}\:.$$
 	\end{Lemma}
\subsection{Related Lemmas}
In this section we will state and prove the related lemmas that are used for proving inequality~\eqref{eq:betti_sum} before proving Theorem~\ref{th:poly} and Theorem~\ref{th:relu}. Lemma~\ref{lemma:two_set_sum} is used for binary classification cases, Lemma~\ref{lemma:complex_isomorphism} and~\ref{lemma:complex_exact} are used for $n$-classification networks.
	\begin{Lemma}\label{lemma:two_set_sum}
		Let $S_1, S_2$ be two sets, Then
		$$b_i(S_1) + b_i(S_2) \leq b_i(S_1 \cup S_2) + b_i(S_1 \cap S_2)$$
		$$b_i(S_1 \cup S_2) \leq b_i(S_1) + b_i(S_2) + b_{i-1}(S_1 \cap S_2)$$
		$$b_i(S_1 \cap S_2) \leq b_i(S_1) + b_i(S_2) + b_{i-1}(S_1 \cup S_2)$$
		In particular, if the homology of $S_1 \cup S_2$ is trivial, then
		$$b_i(S_1) + b_i(S_2) = b_i(S_1 \cap S_2)\:.$$	
	\end{Lemma}

	\begin{proof}
	Recalling the classical Mayer-Vietoris sequence, Let $S_1, S_2\subset \mathbb{R}^n$ be two sets, then the Mayer-Vietoris sequence is the following exact sequence of reduced homology groups
	$$\cdots \rightarrow \widetilde{H}_{i+1}(S_1 \cup S_2) \rightarrow \widetilde{H}_i(S_1 \cap S_2) \rightarrow \widetilde{H}_i(S_1) \bigoplus \widetilde{H}_i(S_2) \rightarrow \widetilde{H}_i(S_1 \cup S_2) \rightarrow \cdots\:,$$
	where $\widetilde{H}_i(S)$ are the reduced homology groups. The results are easy to get.\par
	Suppose that reduced homology groups of $S_1\cup S_2$ are trivial, then 
	$$\widetilde{H}_i(S_1 \cap S_2)\cong \widetilde{H}_i(S_1) \bigoplus \widetilde{H}_i(S_2)$$
	means the equality $b_i(S_1) + b_i(S_2) = b_i(S_1 \cap S_2)$.
	\end{proof}
\begin{Lemma} \label{lemma:complex_isomorphism}
		Supposed that $\phi: \{C^{\bullet}, d\}\rightarrow \{E^{\bullet, \bullet}, d_h, d_v\}$ is a resolution of the single complex $\{C^{\bullet}, d\}$ by a double complex $\{E^{\bullet, \bullet}, d_h, d_v\}$ , Then the induced map 
		$$\phi_{*}: H^{\bullet}(C)\rightarrow H^{\bullet}(T(E))$$
		is an isomorphism. 
\end{Lemma}
	\begin{proof}
	First recall the co-cycle and co-boundary. If the element $(x_{p,q})_{p+q= n} \in T^n(E) = \bigoplus_{p+q=n} E^{p, q}$ satisfies the condition
    $$d_h x_{p -1,q} + d_v x_{p,q} = 0,  \forall p+q=n\:,$$ 
    then it is a co-cycle. If exist the element $(y_{p,q})_{p+q= n- 1} \in T^{n - 1}(E) $ such that $x_{p,q} = d_h y_{p-1, q} +d_v y_{p, q-1}  \forall p+q = n$, then $(x_{p,q})_{p+q= n}$ is a co-boundary. As showed in the zig-zag figure
	$$\xymatrix{
		  	             &         				 &                       &            &       \\   
  x_{p-1,q+1}\ar[r]\ar[u]&				         &                       &            &       \\
		                 & x_{p, q} \ar[u] \ar[r]&                       &            &       \\
		                 &                       &x_{p+1,q-1}\ar[u]\ar[r]&            &       \\
		                 &  		             &  	                 & \ar[u]  	  &
	}
	$$
	~\\ 
	\\ $\phi_{*}$  is surjective: the element $(x_{p,q})_{p+q= n} \in H^n(E)$  is a co-cycle representing the co-homology class $[(x_{p,q})_{p+q= n}]$ that satisfying the condition $D \bigoplus x_{p,q} = 0$. We observe the co-cycle element $(y_{p,q})_{p+q = n}$, $y_{p,q}=0 , \forall q>0$ and $D(y_{p, q}) = 0$, then we follows that 
	$$ d_h y_{n-1, 1} + d_v y_{n, 0} = d_v y_{n, 0} = 0\:. $$ 
	Due to the exactness of the resolution, there exists $u \in C^n$ such that $y_{n,0} = \phi(u)$. Then because of the commute diagram, we have
	$$ 0 = d_h y_{n,0} = d_h \phi(u) = \phi(du)\:. $$
	Since $\phi$ is injective, so 
	$$y_{n,0} = \phi(u), u \in C^n, du=0$$ 
	for the element $(y_{p, q})_{p+q=n}$, a co-cycle $u$ in $ C^n$ can be found. We will inductively prove the co-cycle  $(x_{p,q})_{p+q=n}$ is co-homological with $(y_{p,q})_{p+q=n}, y_{p,q}=0 , \forall q>0$ . 
	\\ $(x_{p,q})_{p+q=n} $ is automatically co-homological with the element $(y^0_{p,q})_{p+q=n}, y^0_{p,q}=0 , \forall p<0$ 
	\\ Supposed that $(x_{p,q})_{p+q=n}$ is co-homological with the   $(y^k_{p,q})_{p+q=n}, y^k_{p,q}=0 , \forall p<k$. Since the exactness of the resolution, $d_v y^{k}_{k, n-k} = 0$, there exists $y_k \text{ s.t } d_v y_k =  y^k_{k, n-k} $, let  $z = (z_{p,q})_{p +q = n-1}\in T^{n-1}(E)$ 
		\begin{equation*}
	    z_{p,q}=\left\{
		\begin{aligned}
		& y_k  &  p=k \\
		& 0    &  else
		\end{aligned}
		\right.\:,
		\end{equation*}
	which implies that the zig-zag $(y^{k+1}_{p,q}) = (y^{k}_{p,q} - D z_{p,q}), y^{k+1}_{p,q}=0, \forall p < k+1$ is co-homological with $(x_{p,q})_{p+q=n}$ 
	~\\
	\\ $\phi_{*}$  is injective. Let $u \in C^n, du = 0$, then the $\phi(u))$ is a co-boundary as showed in the following zig-zag figure,
	$$\xymatrix{
		                         &         				 &                       &            &       \\   
		0 \ar[r]\ar[u]           &				         &                       &            &       \\
		                         & 0 \ar[u] \ar[r]       &                       &            &       \\
		                         &                       &\phi(u)\ar[u]\ar[r]    &            &       \\
		                         &  		             & u \ar[u]^{\phi} 	     &   	      &
	}\:.
	$$
	 Thus there exists $y \in T^{n- 1}(E)$ such that $Dy = \phi(u)$. What we need to prove is to find an element $v \in C^{n-1}$ s.t $u = dv$. Observe the fact that if $y_{p,q}= 0, \forall q > 0$, there exists an element $v \in  C^{n -1}$ such that $\phi(v)= y_{n,0}$, then $\phi(u)  = d_h y_{n,0} = d_h \phi(v) = \phi(dv)$, since the commute diagram and the injection of $\phi, u= dv$, which is the result we want. The only work left is to prove $(y_{p,q})_{p+q = n - 1} \in T^{n-1}(E)$ can be equivalent to another element $(y^{n-2}_{p,q})_{p+q= n-1},y^{n-2}_{p,q} = 0, \forall p < n - 2$ s.t $Dy^{n-2}_{n-1, 0} = \phi(u)$. We inductively to showed this fact using the same trick above. Obviously $y \in T^{n- 1}(E)$ satisfies the condition that $y_{p,q} = 0, \forall p < 0$ automatically. Supposed that $y$ is true under the circumstances that $y_{p,q} = 0, \forall p < k - 1$, there exist an element $(z_{p,q})_{p+q = n-2} \in T^{n -2}(E)$ such that $d_v z_{k-2, n-k} = y_{k-2, n-k+1}$.
	\begin{equation*}
	z_{p,q}=\left\{
	\begin{aligned}
	& z_{k-2, n-k}  &  p=k - 2 \\
	& 0    &  else
	\end{aligned}
	\right.\:.
	\end{equation*}
	Let $y^{k}_{p,q}$ be $ y_{p,q} - Dz_{p,q}$, which satisfies the situation that $(y^{k}_{p,q})_{p+q= n-1},y^{k}_{p,q} =0, \forall p < k $ and $Dy^{k} = \phi(u)$. 
	\end{proof}
\begin{Lemma} \label{lemma:complex_exact}
		The complex sequence
		$$0 \rightarrow C^{*}(S) \stackrel{r}{\longrightarrow} \bigoplus_{\alpha_0} C^{*}(S_{\alpha_0}) \stackrel{\delta}{\longrightarrow} \bigoplus_{\alpha_0< \alpha_1} C^{*}(S_{\alpha_0\alpha_1}) \stackrel{\delta}{\longrightarrow}  \cdots\stackrel{\delta}{\longrightarrow} \bigoplus_{\alpha_0<\cdots < \alpha_p}C^{*}(S_{\alpha_0\cdots\alpha_p})\stackrel{\delta}{\longrightarrow} \cdots $$
		is exact, where $r$ is induced by restriction and the connecting homomorphisms $\delta$ are described above.
\end{Lemma}

	\begin{proof} 
	We set $C^{*}(S_{-1}) = 0$ and $C^{*}(S_0) = C^{*}(S)$, then we can uniformly regard $r = \delta$. First prove that $\delta^2 = 0 $
		\begin{equation*}
			\begin{aligned}
	 	 \delta^2 (\omega)_{\alpha_0\alpha_1\cdots\alpha_{p+2}} &= \sum_{0 \leq i \leq p+2} (-1)^i (\delta \omega)_{\alpha_0\cdots \hat{\alpha}_i\cdots\alpha_{p+2}} \\
	 	 	&= \sum_{0 \leq i \leq p+2} \sum_{j < i} (-1)^i (-1)^j\omega_{\alpha_0\cdots \hat{\alpha}_j\cdots \hat{\alpha}_i\cdots\alpha_{p+2}} \\
	 	  &+\sum_{0 \leq i \leq p+2} \sum_{j > i} (-1)^i (-1)^{j-1}\omega_{\alpha_0\cdots \hat{\alpha}_i\cdots \hat{\alpha}_j\cdots\alpha_{p+2}} \\
	 	 	&=0 \:.
			\end{aligned}
		\end{equation*}
	 	Next let $\omega \in \bigoplus_{\alpha_0<\alpha_1<\cdots < \alpha_p}C^{*}(S_{\alpha_0\alpha_1\cdots\alpha_p})$ such that $\delta\omega = 0$. Write down it precisely, we got
	 	\begin{equation*}
	 	\begin{aligned}
	 	\delta(\omega)_{\alpha\alpha_0\alpha_1\cdots\alpha_{p}} & = \omega_{\alpha_0\alpha_1\cdots\alpha_{p}}  - \sum_{0 \leq i \leq p} (-1)^i \omega_{\alpha\alpha_0\cdots \hat{\alpha}_i\cdots\alpha_{p}} \\
	 	&=0 \:.
	 	\end{aligned}
	 	\end{equation*} 
	 	Hence $\omega_{\alpha_0\alpha_1\cdots\alpha_{p}}  = \sum_{0 \leq i \leq p} (-1)^i \omega_{\alpha\alpha_0\cdots \hat{\alpha}_i\cdots\alpha_{p}}$
	 	Define $\tau \in \bigoplus_{\alpha_0<\alpha_1<\cdots < \alpha_{p-1}}C^{*}(S_{\alpha_0\alpha_1\cdots\alpha_{p-1}})$ by
	 	$$\tau_{\alpha_0\alpha_1\cdots\alpha_{p-1}} = \frac{{1}}{\norm{\{\alpha;\alpha < \alpha_0 < \alpha_1 < \cdots < \alpha_{p-1}\}}} \sum_{\alpha < \alpha_0 < \alpha_1 < \cdots < \alpha_{p-1}}\omega_{\alpha\alpha_0\alpha_1\cdots\alpha_{p-1}}\:.$$
	 	Then
	 	\begin{equation*}
	 		\begin{aligned}
	 		   \delta(\tau)_{\alpha_0\alpha_1\cdots\alpha_{p}} & = \sum_{0 \leq i \leq p} (-1)^i \tau_{\alpha_0\cdots \hat{\alpha}_i\cdots\alpha_{p}} \\
	 		      &= \sum_{0 \leq i \leq p} (-1)^i \frac{{1}}{\norm{\{\alpha;\alpha < \alpha_0  < \cdots < \hat{\alpha}_i < \cdots < \alpha_{p}\}}} \sum_{\alpha < \alpha_0 < \cdots < \hat{\alpha}_i < \cdots < \alpha_{p}}\omega_{\alpha\alpha_0\cdots\hat{\alpha}_i\cdots\alpha_{p}} \\
	 		      &= \omega_{\alpha_0\alpha_1\cdots\alpha_{p}} \:.
	 		\end{aligned}
	 	\end{equation*} 
	 	This gives the wanted result. 
	 	\end{proof}
	\begin{Lemma}\label{lemma:betti_num_sum}
		Let $S_1, S_2,..., S_n$ be the covering of a set $S$, then we have the following inequality 
		$$ b_k(S) \leq \sum_{i + j = k}\sum_{J\subset  \{1,2,..,n\}, \norm{J}=j + 1} b_i(S_J).$$
	\end{Lemma}
	\begin{proof}
	 That is the direct result applied with the Lemma \ref{lemma:complex_isomorphism} and Lemma \ref{lemma:complex_exact}. 
	with the double complex $E^{p, q} = C^{q}(S_{\alpha_0\cdots\alpha_p})$ we have the relation $H_d(C^{*}(S)) \cong H_D(T(E))$, where $D= \delta + (-1)^pd$, we apply the functor $H_d$ to the double complex, so according to the spectral theory, the inequality 
	 $$\text{rank}(H_D(T(E))) \leq \text{rank}(H_d((T(E)))$$ holds, which is the desired result.
	\end{proof}
\subsection{Proofs of Main Theorems}
\begin{proof}
As mentioned above, Lemma~\ref{lemma:polynomial} and Lemma~\ref{lemma:relu} imply decomposition. Apply Lemma ~\ref{lemma:two_set_sum} for binary classification case or Lemma~\ref{lemma:polynomial} and Lemma~\ref{lemma:relu} for $n$-classification cases. Since $\partial S^i(j) = \overline{S^{i}(j)} \cap \overline{U/S^{i}(j)}$, and $U =\mathbb{R}^{n_i}$ or $U =\mathbb{R}_{+}^{n_i}$. Then $U$ is trivial, i.e., $\widetilde{H}(\overline{S^{i}(j)} \cap \overline{U/S^{i}(j)}) \cong \widetilde{H}(\overline{S^{i}(j)}) \bigoplus \widetilde{H}(\overline{U/S^{i}(j)})$, Use Lemma ~\ref{lemma:two_set_sum} 
	combined with Lemma~\ref{lemma:betti_num_sum}, then $b_k(\partial S^i(j)) = b_k(\overline{S^{i}(j)} \cap \overline{U/S^{i}(j)}) = b_k(\overline{S^{i}(j)}) + b_k(\overline{U/S^{i}(j)})$. Thus,
	$$b_k(\overline{S^{i}(j)}) \leq b_k(\partial S^i(j)) \leq \sum_{p + q = k}\sum_{Q\subset \{1,2,..,n\}, \norm{Q}= q + 1} b_p(S_Q^i(j)).$$ From Lemma~\ref{lemma:polynomial} and Lemma~\ref{lemma:relu}, the set $S_{\alpha_0\cdots\alpha_{q}}^i(j)$ is a semi-algebraic set. Apply Lemma ~\ref{lemma:Basu} on the $i$-th layer. When polynomial activation is applied, note that if all $n-1$ component functions are equal, the semi-algebraic set will indeed be an algebraic variety. Use the Lemma~\ref{lemma:Oleinik} in addition, then
	\begin{equation*}
	b_k(\overline{S^{i}(j)})  \\ 
	\leq \left\{
	\begin{aligned}
    &	\sum_{p =0}^{k} \binom{n-1}{p + 1} (3^{n-2-p} - 1) r(l - i - 1) (2r(l - i - 1) - 1)^{n_i - 1},& k < n-2 \\	
	&   \left(\sum_{p =0}^{n-3} \binom{n-1}{p + 1} (3^{n-2-p} - 1) + 1\right) r(l - i - 1) (2r(l - i - 1) - 1)^{n_i - 1},&k \geq n-2
	\end{aligned}
	\right.
	\end{equation*}	
	where $i <  l -1$. In particular, notice that the semi-algebraic set $S^{l-1}(j)$ is constrained by a linear system, then:
	\begin{equation*}
	b_k(\overline{S^{l-1}(j)})  \\ 
	\leq \left\{
	\begin{aligned}
	&	\sum_{p =0}^{k} \binom{n-1}{p + 1} (3^{n-2-p} - 1),      & k < n-2 \\	
	&   \sum_{p =0}^{n-3} \binom{n-1}{p + 1} (3^{n-2-p} - 1) + 1,&k \geq n-2
	\end{aligned}
	\right.
	\end{equation*}		
Similarly, in the Relu case, the covering is still semi-algebraic. Then the following results hold
	\begin{equation*}
	b_k(\overline{S^{i}(j)})  \\ 
	\leq \left\{
	\begin{aligned}
	&	\sum_{p =0}^{k} \binom{n-1}{p + 1} (3^{\sum_{q = l-1}^{i} n_q + n-2-p} - 1)       ,& k < n-2 \\	
	&   \sum_{p =0}^{n-2} \binom{n-1}{p + 1} (3^{\sum_{q = l-1}^{i} n_q + n-2-p} - 1) ,& k \geq n-2.
	\end{aligned}
	\right.
	\end{equation*}	
\end{proof}
\newpage
\section{Supplementary Material}
\subsection{Related Definitions}
\begin{definition}[Simplices]
	The \textit{n-complex}, $\Delta^n$, is the simplest geometric convex hull spanned by a collection of $n + 1$ points in Euclidean space $\mathbb{R}^{n+1}$, generally denoted as $[e_0, e_1, ..., e_n]$, where $e_i = (0,..,0,1,0,..0)$ the i-th position is $1$ and other else is $0$.
\end{definition}

\begin{definition}[Simplicial Complexes]
	A simplicial complex $X$ is a finite set of simplices satisfying the following conditions:
	\begin{enumerate}
	    \item For all simplices $\sigma \in X$, $\alpha$ is a face of  $\sigma$ and $\alpha \in X$.
	    \item $\sigma_1, \sigma_2 \in X$ such that $\sigma_1, \sigma_2$ are properly situated.
	\end{enumerate}
\end{definition}

\begin{definition}[Connected and path-connected]
	A complex $X$ is connected if it can not be represented as the disjoint union of two or more non-empty sub-complexes. A geometric complex is path-connected if there exists a path made of $1$-simplices.
\end{definition}

\begin{definition}[n-chain]
	Given a set of n-simplices $\sigma_1, \sigma_2, ...,\sigma_k$ in a complex and an Abelian group $G$, we define an n-chain with coefficients in $G$ as a formal sum
	$$x = \sum_{i} g_i \cdot \sigma_i\:,$$
	where $g_i \in G$. The set of all n-chains equipped with additive operator forms a new Abelian group, n-chain group denoted as $C_n(X)$.
\end{definition}

\begin{definition}[boundary operation]
	Let $\sigma$ be an oriented n-simplex in a complex $X$. The boundary of it is defined as the $(n-1)$-chain of $X$ over $\mathbb{Z}$ given by
	$$\delta_n : C_n(X)\rightarrow C_{n-1}(X)$$
	is a homomorphism from n-chain group to $(n-1)$-chain group.
	$$\delta_n \sigma = \sum_{i=0}^n (-1)^i [v_0,...,v_{i-1}, v_{i+1},...,v_n].$$ 
\end{definition}
Then we can easily verify that $\delta_{n+1} \delta_n \sigma = 0$, where $\sigma$ is a $n+1$-simplex. It shows the fact that 
$$C_{n+1}(X) \stackrel{\delta_{n+1}}{\longrightarrow} C_n(X) \stackrel{\delta_{n}}{\longrightarrow}  C_{n-1}(X).$$
i.e. $Img(\delta_{n+1})\subset Ker(\delta_{n}) \subset C_n(X)$.

\begin{definition}[Cohomology group]
   Given a chain complex  \[\cdots {\rightarrow} C_{n+1}(X) \stackrel{\delta_{n+1}}{\longrightarrow} C_n(X) \stackrel{\delta_{n}}{\longrightarrow}  C_{n-1}(X)\to\dots\to C_1(X)\stackrel{\delta_{1}}{\longrightarrow}C_0(X)\stackrel{\delta_{0}}{\longrightarrow}0 {\longrightarrow}\cdots\] 
   and a group $G$, we can define the co-chains $C^{*}_n$ to be the respective groups of all  homomorphsims from $C_n$ to $G$:
   $$C^{*}_n = \text{Hom}(C_n, G)$$
    We define the co-boundary map $d_n: C_{n}^{*} \rightarrow C_{n-1}^{*}$ dual to $\delta_n$ as the map sending $\phi \mapsto d_n \phi = \delta^{*}\phi$. For an element $c \in C_n$ and a homomorphism $\phi \in C_{n-1}^{*}$, we have $$d_n \phi(c) = \phi(\delta_n c)$$
    Because $\delta_n\delta_{n+1}=0$, it is easily seen  that $d_{n+1}d_n =0$. In other words, $\text{Img } d_n \subset \text{Ker } d_{n+1}$. With this fact we can define the $n$-th cohomology group as the quotient:
    $$H^n(C; G) =  \text{Ker } d_{n+1} / \text{Img } d_n$$
\end{definition}

\begin{definition}[Reduced homology groups]
   the augmented chain complex is defined as \[\cdots {\rightarrow} C_{n+1}(X) \stackrel{\delta_{n+1}}{\longrightarrow} C_n(X) \stackrel{\delta_{n}}{\longrightarrow}  C_{n-1}(X)\to\dots\to C_1(X)\stackrel{\delta_{1}}{\longrightarrow}C_0(X)\stackrel{\epsilon}{\longrightarrow}\sZ\stackrel{\delta_{0}}{\longrightarrow}0 {\rightarrow}\cdots\]
where $\epsilon(\sum_i n_i\sigma_i) = \sum_i n_i$, then
the reduced homology groups $\widetilde{H}_i(X) = \Ker(\delta_i)/\Img(\delta_{i+1})$ for $i>0$, and $\widetilde{H}_0(X) = \ker(\epsilon)/\Img(\delta_1)$. One can show that $H_i(X) = \widetilde{H}_i(X)$ for $i>0$ and $H_0(X) = \widetilde{H}_0(X) \bigoplus \sZ$.
\end{definition}

\begin{definition}[exact sequence]
	A sequence 
	$$\cdots \rightarrow G_0  \stackrel{f_0}{\longrightarrow} G_1 \stackrel{f_1}{\longrightarrow} G_2 \stackrel{f_2}{\longrightarrow}  \cdots$$
	of groups and group homomorphisms is called exact if the image of  each homomorphism  is equal to the kernel of the next:
	$$Img(f_k) = Ker(f_{k+1})$$
	the sequence of groups and homomorphism may be either finite or infinite.
\end{definition}

	\begin{definition}[Double complex]
		A double complex  is a bi-graded Abelian groups 
		$$E^{\bullet, \bullet} = \bigoplus_{p, q} E^{p,q} $$
		equipped with the two homomorphisms
			$$ d_h : E^{\bullet, \bullet} \rightarrow E^{\bullet,\bullet+1} , \quad d_v :E^{\bullet, \bullet} \rightarrow E^{\bullet+1 ,\bullet}$$
		satisfying the conditions:
			$$d_h^2 =  d_v^2 = d_vd_h + d_hd_v=0$$
	\end{definition} 
	the above equations give the $D^2 = 0 $ if we define $D = d_h +d_v$ and without generality, let $$E^{p,q}=0, p <0 \text{ or } q <0$$ and a double complex can induce a single complex
	\begin{definition}[total complex]
		The total complex associated to a double complex $\{E^{\bullet,\bullet}, d_h, d_v\}$ is the complex $\{T^{\bullet}(E), D\}$ where 
		$$T^n(E) := \bigoplus_{p+q=n}E^{p,q}$$
	\end{definition}
	
	With the double complex associated its total complex, A complex of Abelian groups $\{C^{\bullet}, d\}$ can be approximated under the following commute diagram:
	
	$$\xymatrix{
		               & \vdots                           & \vdots                          & \vdots                          &  \\   
		0 \ar[r]^{d_h} & E^{0,1} \ar[r]^{d_h}\ar[u]^{d_v} & E^{1,1}\ar[r]^{d_h}\ar[u]^{d_v} & E^{2,1}\ar[r]^{d_h}\ar[u]^{d_v} &\cdots \\
		0 \ar[r]^{d_h} & E^{0,0} \ar[r]^{d_h}\ar[u]^{d_v} & E^{1,0}\ar[r]^{d_h}\ar[u]^{d_v} & E^{2,0}\ar[r]^{d_h}\ar[u]^{d_v} &\cdots \\
		0 \ar[r]^{d}   & C^{0}   \ar[r]^{d}  \ar[u]^{\phi}& C^{1}  \ar[r]^{d}  \ar[u]^{\phi}& C^{2}  \ar[r]^{d}  \ar[u]^{\phi}&\cdots \\
		               & 0 		             \ar[u]       & 0 		             \ar[u]     & 0 		             \ar[u]   &
	}
	$$
	In particular, if we have the condition:
	\begin{definition}[resolution]
		A resolution of the complex $\{C^{\bullet},d\}$ by a double complex$\{E^{\bullet, \bullet}, d_h, d_v\}$ is a homomorphism $\phi: \{C^{\bullet}, d\}\rightarrow \{E^{\bullet, \bullet}, d_h, d_v\}$ such that the columns in the above diagram are exact. In another word, we have the long exact sequence:
		$$0\hookrightarrow C^n \stackrel{\phi}{\longrightarrow} E^{n,0} \stackrel{d_v}{\longrightarrow} E^{n,1} \stackrel{d_v}{\longrightarrow} \cdots$$
	\end{definition}

\subsection{Related Property}
\begin{Property}
	If $X_1, ..., X_p$ is the set of all connected components of a complex $X$, and $H_n, H_n^i$ are the homology groups of $X$ and $X_i$ respectively, then $H_n$ is isomorphic to the direct sum $H_n^1\bigoplus \cdots \bigoplus H_n^p$.
\end{Property}
\begin{Property}
	The zero-dimensional homology group of a complex $X$ over $\mathbb{Z}$ is isomorphic to $\mathbb{Z}^p= \bigoplus_p \mathbb{Z}$, where $p$ is  the number of connected components of $X$. 
\end{Property}
The above two properties show topological intuition on the connected components of the dataset. Moreover, the higher-dimensional homology groups show the "higher dimensional holes" of the topological space. From the definition of the homology groups, it is clear that when the dimensions of the homology groups are no less than the dimension of topological space, it will degenerate to a trivial group, i.e., $\{0\}$. 

\subsection{Related Theorems}
The two related theorems from~\cite{Basu2001}.
\begin{Theorem}
	Let $\mathbf{R}$ be a real closed field and let $S \subset \mathbf{R}^k$ be the set define by the conjunction of $n$ inequalities.
	$$\mathbf{P}_1 \geq 0, \cdots, \mathbf{P}_n \geq 0$$
	$\mathbf{P}\in \mathbf{R}[X_1,...,X_k], deg(\mathbf{P}_i) \leq d, 1\leq i\leq n$, contained in a variety $\mathbf{Z(Q)}$ of real dimension $k'$ with $deg(\mathbf{Q}) \leq d$. Then we have
	$$b_i(S) \leq \sum_{j=0}^{k'-i} \binom{n}{j} 2^{j+1} d(2d - 1)^{k- 1} \leq \binom{n}{k'-i} \mathbf{O}(d^k).$$	
\end{Theorem}
and its dual result:
\begin{Theorem}
	Let $\mathbf{R}$ be a real closed field and let $S \subset \mathbf{R}^k$ be the set define by the disjunction of $n$ inequalities.
	$$\mathbf{P}_1 \geq 0, \cdots, \mathbf{P}_n \geq 0$$
	$\mathbf{P}\in \mathbf{R}[X_1,...,X_k], deg(\mathbf{P}_i) \leq d, 1\leq i\leq n$. Then we have
	$$b_i(S) \leq \sum_{j=0}^{i+1} \binom{n}{j} 3^{j} d(2d - 1)^{k- 1} \leq \binom{n}{i+1} \mathbf{O}(d^k).$$	
\end{Theorem}

\subsection{More Experimental results}
\begin{figure*}[!ht]
  \centering
  \setlength\tabcolsep{2pt} 
  \def\arraystretch{1} 
  \newcommand{\mywidth}{0.3\textwidth} 
  \newcommand{\mywidthlr}{0.05\textwidth} 
  \newcommand{\mywidths}{0.45\textwidth}
  \newcolumntype{R}{>{\centering\arraybackslash}m{\mywidth}}
  \newcolumntype{Y}{>{\centering\arraybackslash}m{\mywidthlr}}
\begin{tabular}{YRRR}
\rotatebox[origin=c]{90}{layer 1} &
\includegraphics[width = \mywidth]{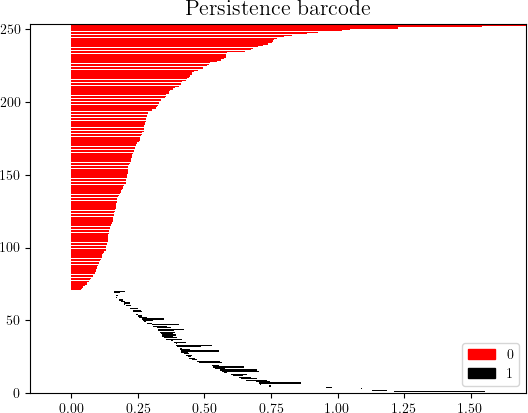}&
\includegraphics[width = \mywidth]{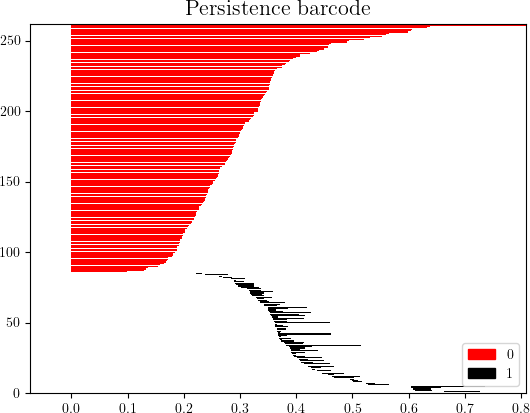}&
\includegraphics[width = \mywidth]{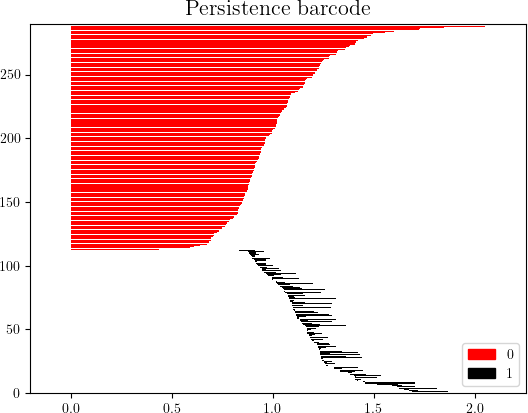} \\
\rotatebox[origin=c]{90}{layer 2} &
\includegraphics[width = \mywidth]{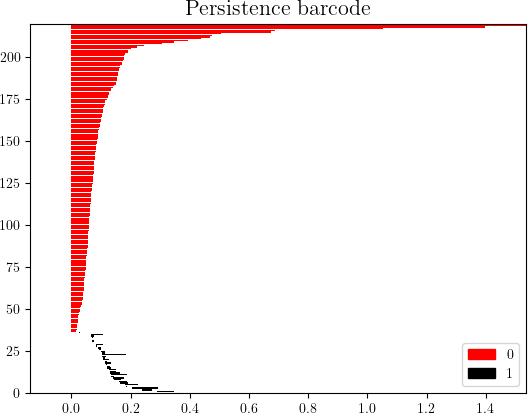}&
\includegraphics[width = \mywidth]{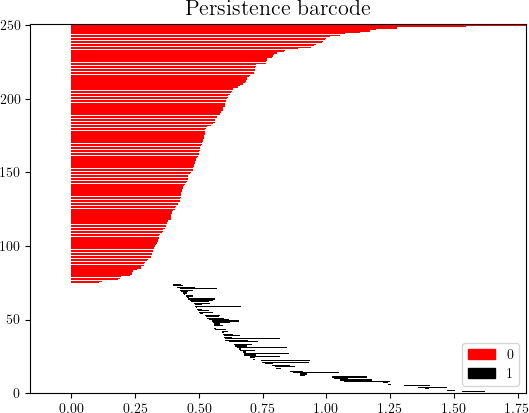}&
\includegraphics[width = \mywidth]{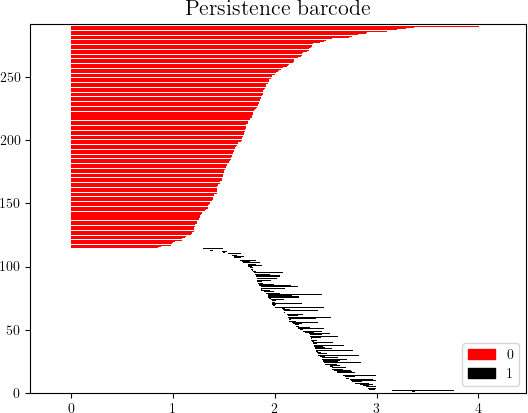} \\
\rotatebox[origin=c]{90}{layer 3} &
\includegraphics[width = \mywidth]{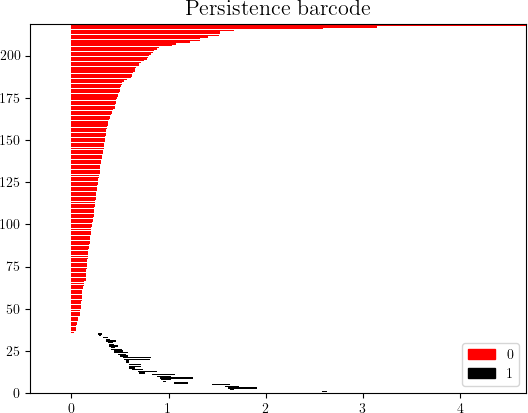}&
\includegraphics[width = \mywidth]{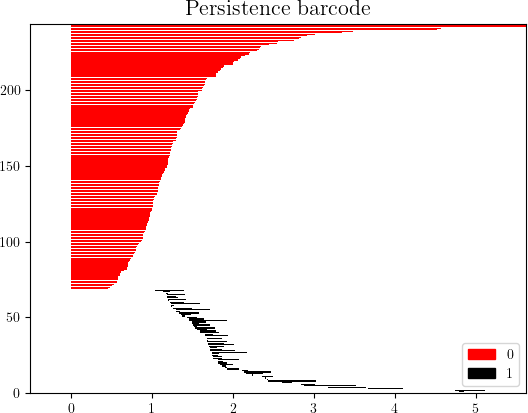}&
\includegraphics[width= \mywidth]{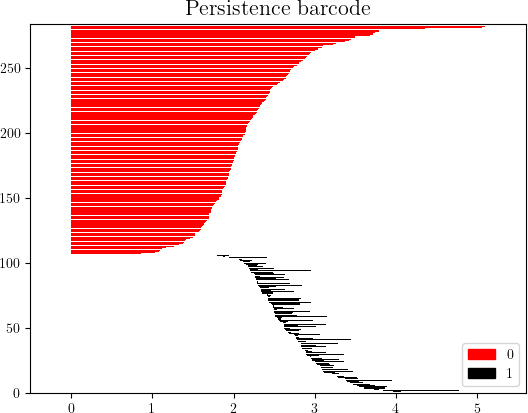}\\
& node = 4 & node = 10 & node = 26 \\
\end{tabular}
  \caption{The graphs shows the different persistent barcodes graphs on each layer (with ReLU activation) with variant hidden units number from class 0. Note that the expressivity decrease along each column and increase along each row. The figure corroborates our results that the Betti numbers decrease along layers and increase when node number increases.}
 \label{fig:layer0_layer_node}
\end{figure*}

\begin{figure*}[!ht]
  \centering
  \setlength\tabcolsep{2pt} 
  \def\arraystretch{1} 
  \newcommand{\mywidth}{0.3\textwidth} 
  \newcommand{\mywidthlr}{0.05\textwidth} 
  \newcommand{\mywidths}{0.45\textwidth}
  \newcolumntype{U}{>{\centering\arraybackslash}m{\mywidth}}
  \newcolumntype{Y}{>{\centering\arraybackslash}m{\mywidthlr}}
\begin{tabular}{YUUU}
\rotatebox[origin=c]{90}{layer 1} &
\includegraphics[width = \mywidth]{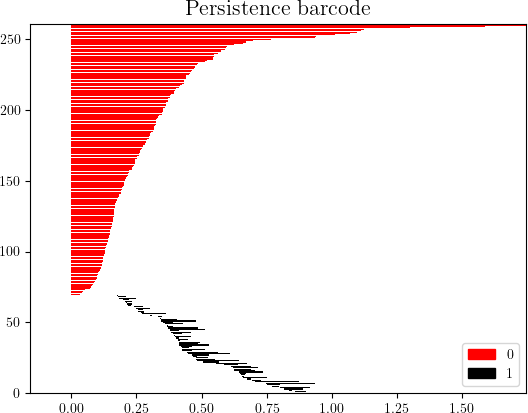}&
\includegraphics[width = \mywidth]{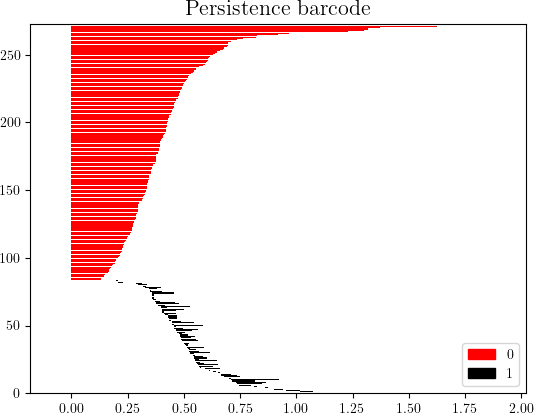}&
\includegraphics[width = \mywidth]{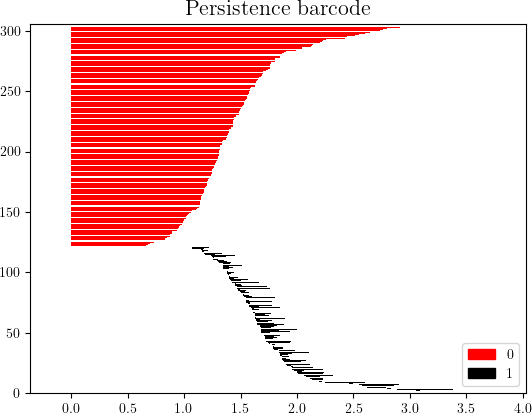} \\
\rotatebox[origin=c]{90}{layer 2} &
\includegraphics[width = \mywidth]{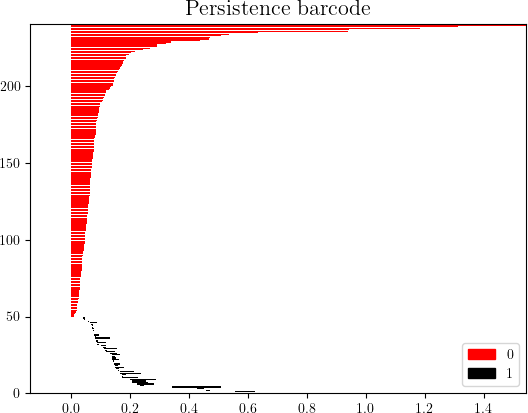}&
\includegraphics[width = \mywidth]{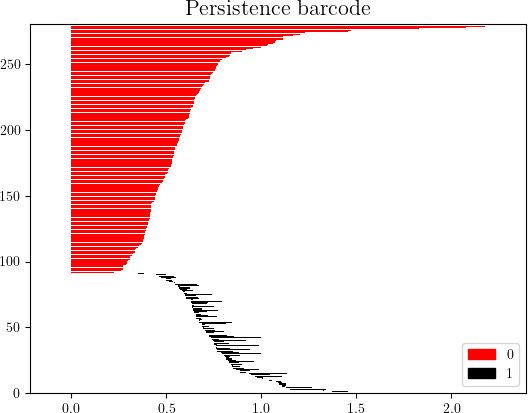}&
\includegraphics[width = \mywidth]{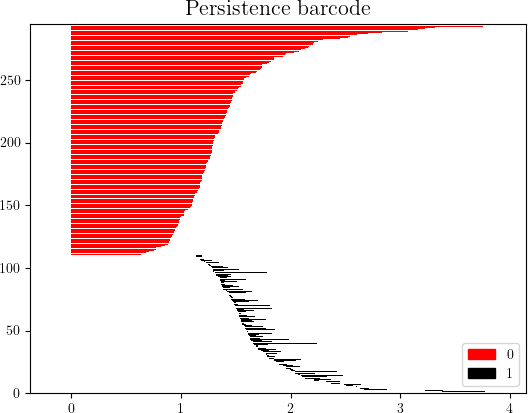} \\
\rotatebox[origin=c]{90}{layer 3} &
\includegraphics[width = \mywidth]{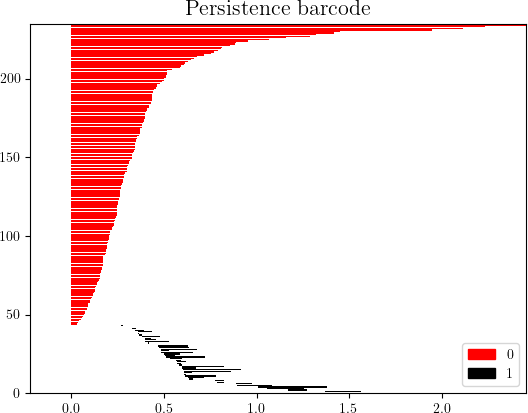}&
\includegraphics[width = \mywidth]{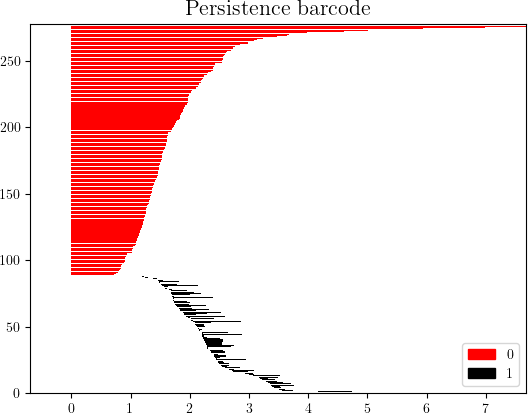}&
\includegraphics[width= \mywidth]{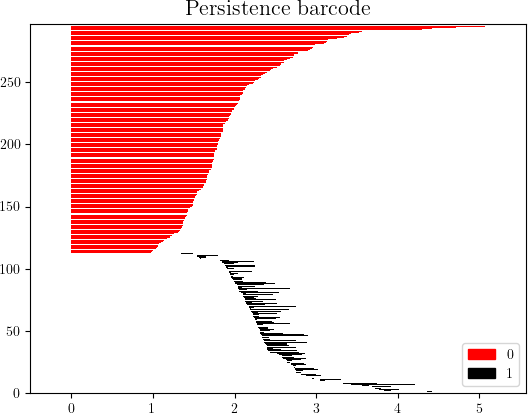}\\
& node = 4 & node = 12 & node = 25 \\
\end{tabular}
  \caption{The graphs shows the different persistent barcodes graphs on each layer (with ReLU activation) with variant hidden units number from class 2. Note that the expressivity decrease along each column and increase along each row. The figure corroborates our results that the Betti numbers decrease along layers and increase when node number increases.}
\end{figure*}

\begin{figure*}[!ht]
  \centering
  \setlength\tabcolsep{2pt} 
  \def\arraystretch{1} 
  \newcommand{\mywidth}{0.48\textwidth} 
  \newcommand{\mywidthlr}{0.25\textwidth} 
  \newcommand{\mywidths}{0.2\textwidth}
  \newcolumntype{T}{>{\centering\arraybackslash}m{\mywidth}}
  \newcolumntype{Y}{>{\centering\arraybackslash}m{\mywidths}}
\begin{tabular}{TT}
\includegraphics[width = \mywidth]{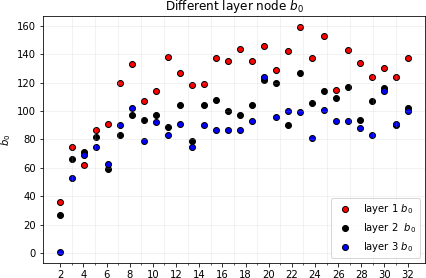} &
\includegraphics[width = \mywidth]{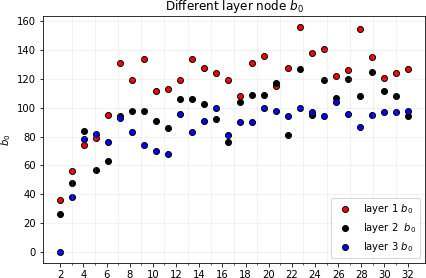} \\
 class 2 &  class 3\\
\includegraphics[width = \mywidth]{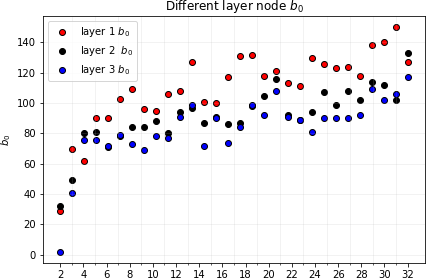} &
\includegraphics[width = \mywidth]{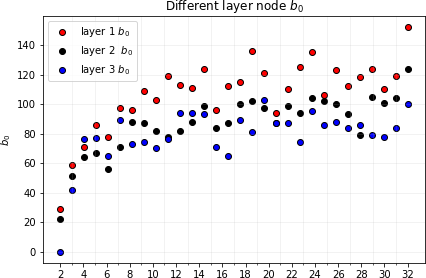} \\
\centering class 4 & \centering class 6
\end{tabular}
  \caption{For a network with three equal-size dense layers (with polynomial activation), change the layer size, Betti numbers of the different layers are descending along the layers.}
  \label{fig:supp_layer_decrease}
  \vspace{3ex}
  \begin{tabular}{TT}
\includegraphics[width = \mywidth]{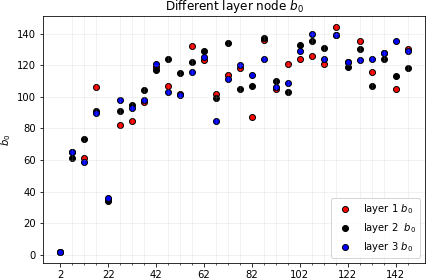} &
\includegraphics[width = \mywidth]{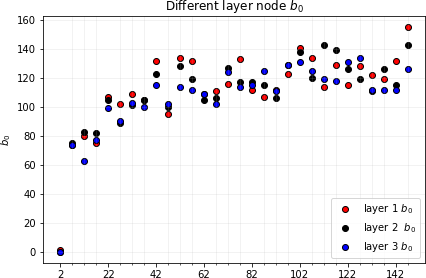} \\
 class 2 &  class 3\\
\includegraphics[width = \mywidth]{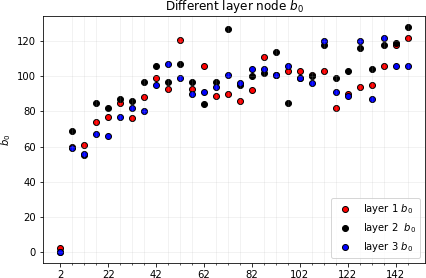} &
\includegraphics[width = \mywidth]{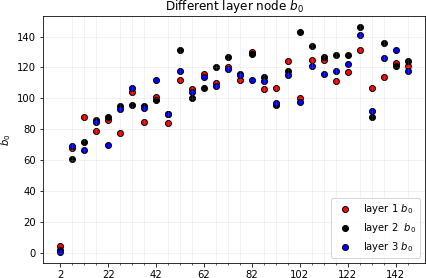} \\
\centering class 5 & \centering class 8
\end{tabular}
  \caption{The homology structures within the layers are preserved if the number of hidden nodes is sufficient and the complexity of the input exceeds the expressiveness of a network with fewer nodes.}
  \label{fig:supp_layer_decrease_relu}
\end{figure*}

\end{document}